%% file: omnilabel_benchmark_arxiv.tex
\documentclass[10pt,twocolumn,letterpaper]{article}

\usepackage{iccv}
\usepackage{times}
\usepackage{epsfig}
\usepackage{graphicx}
\usepackage{amsmath}
\usepackage{amssymb}

\input{preamble1}

\usepackage[breaklinks=true,bookmarks=false]{hyperref}

\input{preamble2}

\iccvfinalcopy 

\def\iccvPaperID{7127} 

\begin{document}

\title{OmniLabel: A Challenging Benchmark for Language-Based Object Detection}

\author{
Samuel Schulter\textsuperscript{1, \textdagger} \quad Vijay Kumar B G\textsuperscript{1} \quad Yumin Suh\textsuperscript{1} \quad Konstantinos M. Dafnis\textsuperscript{2,*} \vspace{0.5em} \\
Zhixing Zhang\textsuperscript{2,*} \quad Shiyu Zhao\textsuperscript{2,*} \quad Dimitris Metaxas\textsuperscript{2} \vspace{0.5em} \\
{\normalsize \textsuperscript{\textdagger} project lead \quad \textsuperscript{*} equal technical contribution, alphabetic order} \vspace{0.3em}\\
\textsuperscript{1} NEC Laboratories America \quad \textsuperscript{2} Rutgers University
}

\maketitle

\input{sections/abstract.tex}
\input{sections/intro.tex}
\input{sections/relatedwork.tex}
\input{sections/benchmark.tex}
\input{sections/datacollection.tex}
\input{sections/dataanalysis.tex}
\input{sections/baselines.tex}

\input{sections/conclusion.tex}

{\small
\bibliographystyle{ieee_fullname}
\bibliography{egbib}
}

\appendix
\clearpage
\input{sections_supp/supp_dataanalysis}

\input{sections_supp/supp_datacollection}

\input{sections_supp/supp_code_and_dataset}
\input{sections_supp/supp_dataset_examples}

\end{document}

%% file: preamble1.tex
\usepackage{booktabs}
\usepackage{pifont}
\newcommand{\cmark}{\ding{51}}
\newcommand{\xmark}{\ding{55}}
\usepackage{enumitem}
\usepackage{caption}
\usepackage{subcaption}
\usepackage{lipsum}
\usepackage{multirow}

%% file: preamble2.tex
\usepackage[capitalize]{cleveref}
\crefname{section}{Sec.}{Secs.}
\Crefname{section}{Section}{Sections}
\Crefname{table}{Table}{Tables}
\crefname{table}{Tab.}{Tabs.}

\makeatletter
\renewcommand{\paragraph}{%
  \@startsection{paragraph}{4}%
  {\z@}{0.5ex \@plus 1ex \@minus .2ex}{-1em}%
  {\normalfont\normalsize\bfseries}%
}
\makeatother

\usepackage{enumitem}
\setitemize[0]{leftmargin=10pt,itemsep=0pt,topsep=2pt,parsep=0pt}
\setenumerate[0]{leftmargin=16pt,itemsep=0pt,topsep=2pt,parsep=0pt}

\usepackage[table]{xcolor}
\definecolor{cmark}{RGB}{50,210,60}
\definecolor{xmark}{RGB}{210,50,60}
\newcommand{\cmarkcol}{\textcolor{cmark}{\cmark}}
\newcommand{\xmarkcol}{\textcolor{xmark}{\xmark}}

%% file: sections/abstract.tex
\begin{abstract}
Language-based object detection is a promising direction towards building a natural interface to describe objects in images that goes far beyond plain category names. While recent methods show great progress in that direction, proper evaluation is lacking. With OmniLabel, we propose a novel task definition, dataset, and evaluation metric. The task subsumes standard- and open-vocabulary detection as well as referring expressions. With more than 28K unique object descriptions on over 25K images, OmniLabel provides a challenging benchmark with diverse and complex object descriptions in a naturally open-vocabulary setting. Moreover, a key differentiation to existing benchmarks is that our object descriptions can refer to one, multiple or even no object, hence, providing negative examples in free-form text. The proposed evaluation handles the large label space and judges performance via a modified average precision metric, which we validate by evaluating strong language-based baselines. OmniLabel indeed provides a challenging test bed for future research on language-based detection. Visit the project website at \url{https://www.omnilabel.org}
\end{abstract}

%% file: sections/intro.tex
\section{Introduction}
\label{sec:intro}
A nuanced understanding of the rich semantics of the world around us is a key ability in the visual perception system of humans. Identifying objects from a description like ``person wearing blue-and-white striped T-shirt standing next to the traffic sign'' feels easy, because humans understand the composition of object category names, attributes, actions, and spatial or semantic relations between objects. When automated, this same ability can improve and enable a plethora of applications in robotics, autonomous vehicles, navigation, retail, etc.

\begin{figure}
    \centering
    \includegraphics[width=1.0\columnwidth]{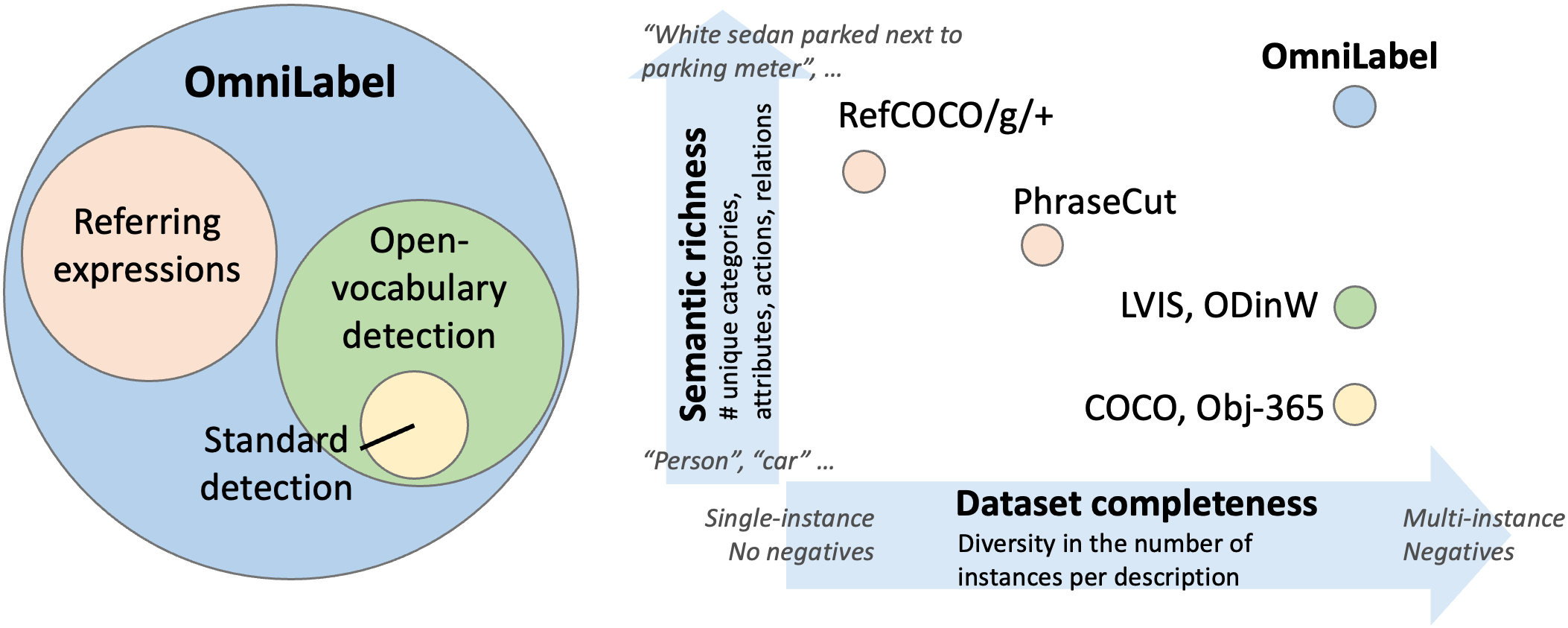}
    \caption{
    \textbf{(Left)} OmniLabel extends upon standard detection, open-vocabulary detection, as well as referring expressions, subsuming these tasks as special cases.
    \textbf{(Right)} The OmniLabel dataset is semantically rich with diverse free-form text descriptions of objects. Moreover, with object descriptions referring to multiple instances, dedicated negative descriptions, and a novel evaluation metric, our benchmark poses a challenging task for language-based detectors.
    }
    \label{fig:teaser}
\end{figure}

With the recent advances in vision \& language models~\cite{jia_icml_21_align,khan_eccv22_simla,radford_icml_2021_CLIP}, along with extensions towards object localization~\cite{gu_iclr_22,kamath2021mdetr,li_iclr_22,li_cvpr22_glip,zhao_eccv22_vlplm}, a comprehensive evaluation benchmark is needed. However, existing ones fall short in various aspects. While object detection datasets significantly increased the label space over time (from 20 in Pascal~\cite{Everingham_2010_IJCV_VOC_PASCAL} to 1200 in LVIS~\cite{gupta_cvpr19_lvis}), a fixed label space is assumed. The zero-shot~\cite{bansal2018zero} and open-vocabulary detection~\cite{gu_iclr_22,zareian_cvpr_21} settings drop the fixed-labelspace assumption, but corresponding benchmarks only evaluate simple category names, neglecting more complex descriptions. Referring expression datasets~\cite{mao_cvpr16_refcocog,yu_eccv16_refcoco} probe models with free-form text descriptions of objects. However, the corresponding dataset annotations and metrics do not allow for a comprehensive evaluation of models.

We introduce a novel benchmark called OmniLabel with the goal to comprehensively probe models for their ability to understand complex, free-form textual descriptions of objects and to locate the corresponding instances. This requires a novel task definition and evaluation metric, which we propose in \cref{sec:benchmark}. Our evaluation benchmark does not assume a fixed label space (unlike standard detection), uses complex object descriptions beyond plain category names (unlike open-vocabulary detection), and evaluates true detection ability with descriptions referring to zero, one or more instances in a given image (unlike referring expressions). A unique aspect of our benchmark are the descriptions that refer to zero instances, which pose a challenge to existing methods as hard negative examples. \cref{fig:teaser} positions our OmniLabel benchmark.

\begingroup
\setlength{\tabcolsep}{3.0pt} 
\renewcommand{\arraystretch}{1.3} 
\begin{table}[t]\centering
    \begin{tabular}{l c c c c c c c c}
    Dataset   & \rotatebox{90}{\# images}  & \rotatebox{90}{Free-form} & \rotatebox{90}{Descr. length} & \rotatebox{90}{\# unique nouns} & \rotatebox{90}{Open-vocabulary} & \rotatebox{90}{Multi-Instance} & \rotatebox{90}{Negative} & \rotatebox{90}{Evaluation} \\
    \toprule
    LVIS~\cite{gupta_cvpr19_lvis}  & 5K    & \xmarkcol & -- & 1.2K & \xmarkcol & \cmarkcol & \cmarkcol & AP \\
    ODinW~\cite{li_22_elevater_odinw}   & 27.3k   & \xmarkcol & -- & 0.3K  & \cmarkcol & \cmarkcol & \cmarkcol & AP \\
    RefCOCO~\cite{mao_cvpr16_refcocog,yu_eccv16_refcoco}     & 4.3K  & \cmarkcol & 4.5 & 3.5K & \cmarkcol & \xmarkcol & \xmarkcol & P \\
    Flickr30k~\cite{flickrentitiesijcv}    & 1.0K  & \cmarkcol & 2.4 & 1.9K & \cmarkcol & \xmarkcol & \xmarkcol & R \\
    PhraseCut~\cite{wu2020phrasecut}  & 2.9K  & \cmarkcol & 2.0 & 1.5K & \cmarkcol & \cmarkcol & \xmarkcol & IoU\\
    \rowcolor{green!5}
    \textbf{OmniLabel}      & 12.2K & \cmarkcol & 5.6 & 4.6K & \cmarkcol & \cmarkcol & \cmarkcol & AP \\
    \bottomrule
    \end{tabular}
    \caption{
    Comparing OmniLabel to existing benchmarks: On 12.2K images, OmniLabel provides free-form text descriptions of objects with an average description length of 5.6 words, covering 4.6K unique nouns. Each description can refer to multiple objects, or no object, \ie, a negative example, an important factor in our evaluation. (Numbers are computed on validation sets. P: Precision, R: Recall, AP: Average Precision, IoU: Intersection over Union)
    }
    \label{tab:dataset_stats_basic}
\end{table}
\endgroup

\begin{figure}\centering
    \includegraphics[width=1.0\columnwidth]{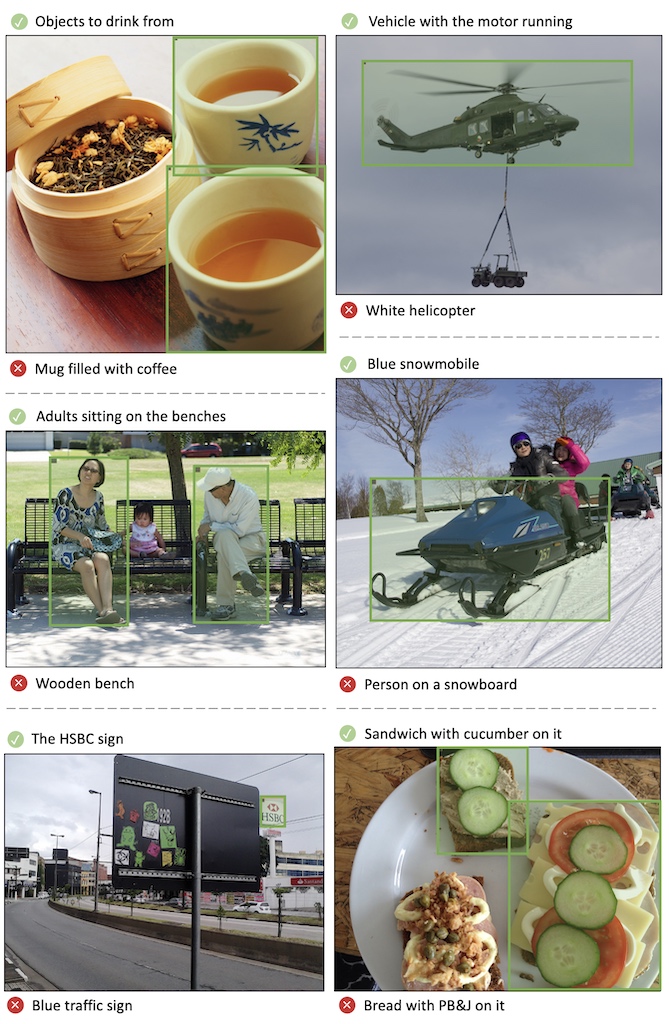}
    \caption{
    Examples of the OmniLabel ground truth annotations. Positive descriptions (above each image) can refer to one or multiple instances. Negative descriptions (below each image) are semantically related to the image but refer to no object.
    More examples in \cref{sec:supp_dataset_examples}
    }
    \label{fig:omnilabel_examples}
\end{figure}

To build this evaluation benchmark, we collected a set of novel annotations upon existing object detection datasets. We augment the existing plain category names with novel free-form text descriptions of objects. Our specific annotation process (\cref{sec:data_collection}) increases the difficulty of the task by ensuring that at least one of the following conditions is true:
\begin{enumerate}[label=(\alph*)]
    \item Multiple instances of the same underlying object category are present in the same image
    \item One object description can refer to multiple objects
    \item An image contains a negative object description, which refers to no object but is related to the image's semantics
    \item Descriptions do not use the original category name
\end{enumerate}
\cref{tab:dataset_stats_basic} highlights the key differences of OmniLabel to existing benchmarks: The diversity in the free-form text descriptions and the evaluation as an object detection task, including multiple instances per description as well as negative object descriptions. The numbers in the table reflect our \emph{public validation set}, which is roughly the same size as our \emph{private test set}. \cref{fig:omnilabel_examples} provides examples of the dataset.

We also evaluate recent language-based detectors on our benchmark, including RegionCLIP~\cite{zhong2022regionclip}, Detic~\cite{zhou2021detic}, MDETR~\cite{kamath2021mdetr}, GLIP~\cite{li_cvpr22_glip} and FIBER~\cite{dou2022coarse}. Summarized in \cref{subsec:baseline_results}, our key observation is that the proposed benchmark is difficult for all methods, and the evaluation metric is more stringent than in prior benchmarks. Negative object descriptions pose the biggest challenge to current methods.

We summarize our contributions as follows:
\begin{enumerate}[label=(\alph*)]
    \item A novel benchmark to unify standard detection, open-vocabulary detection and referring expressions
    \item A data annotation process to collect diverse and complex free-form text descriptions of objects, including negative examples
    \item A comprehensive evaluation metric that handles the virtually infinite label space
\end{enumerate}

%% file: sections/relatedwork.tex
\section{Related Work}
\label{sec:relatedwork}

To position our proposed OmniLabel benchmark, we relate it to existing tasks and focus on the corresponding benchmark datasets.

\paragraph{Object Detection:}
Localizing and categorizing objects is a long-standing and important task with many applications. An enormous amount of datasets fueled this research. Besides datasets for specific use cases, like face~\cite{yang2016widerface}, pedestrians~\cite{andriluka_08_tudped,pdollar_2012_caltechped} or driving scenes~\cite{Cordts2016Cityscapes,Geiger_2012_CVPR_KITTI,neuhold_iccv_2017_vistas,Yu_2018_arxiv_bdd100k}, the most popular ones are general-purpose: Pascal VOC~\cite{Everingham_2010_IJCV_VOC_PASCAL}, MS COCO~\cite{lin_eccv_14_mscoco}, Objects-365~\cite{Objects365}, OpenImages~\cite{OpenImages} or LVIS~\cite{gupta_cvpr19_lvis}. These datasets also reflect the evolution of size, both in number of images and, more relevant here, the number of categories. In the same order as above, the label space sizes are 20, 80, 365, 600 and 1203. These datasets lead to significant progress in the past years on neural network architectures~\cite{carion2020end,lin_2017_fpn,liu2016ssd,redmon_cvpr17_yolo9k,Ren_2015_NIPS,zhang2022_detr_dino,zhou2019objects_centernet,zhu2020deformable_detr} as well as robustness and scaling~\cite{zhao_eccv20_multidatasetdet,zhou2021detic,zhou_cvpr22_simple}. Still, the limitations over OmniLabel are obvious: All detection datasets assume a fixed labelspace, do not provide an open-vocabulary setting or free-form object detections.

\paragraph{Referring Expressions:}
Instead of a limited and fixed set of category names, the motivation in referring expressions is to refer to objects with natural language.
The most popular benchmark is the series of RefCOCO/g/+~\cite{mao_cvpr16_refcocog,yu_eccv16_refcoco}. While RefCOCO/g~\cite{mao_cvpr16_refcocog} often contains long and redundant descriptions, RefCOCO/+~\cite{yu_eccv16_refcoco} limited the referring phrases with a specific annotation process involving a two-player game. The RefCOCO+ extension restricted annotators to use spatial references (\eg, ``man on left''), which was likely over-used because all of RefCOCO/g/+ assume each phrase to refer to exactly one instance. In contrast, OmniLabel explicitly asks annotators to pick two or more instances to describe in many images. PhraseCut~\cite{wu2020phrasecut} also collects templated expressions that refer to multiple instances and also provides segmentation masks. However, OmniLabel still has more instances per object description and uses free-form descriptions. Moreover, none of the existing referring expression datasets provides negative examples.

\paragraph{Visual Grounding:}
While the task of referring expressions is to localize the main subject of the phrase, visual grounding aims at localizing each object of the phrase, \ie, grounding the text in the image. Benchmarks include Flickr30k~\cite{flickrentitiesijcv} or Visual Genomes~\cite{krishna2017visual}, which have often been used also for general object-centric pre-training for vision \& language models like GLIP~\cite{li_cvpr22_glip},  MDETR~\cite{kamath2021mdetr}, SIMLA~\cite{khan_eccv22_simla}, ALBEF~\cite{ALBEF}. 
OmniLabel addresses a different task that is more related to referring expressions. The annotation costs for grounding are also typically higher since all objects mentioned in a phrase need an associated bounding box, which often leads to noisy ground truth. In contrast, the annotation process for OmniLabel can easily be built upon existing detection datasets with high-quality bounding boxes.

\paragraph{Open-Vocabulary Object Detection:}
Aside from using natural language as object descriptions, scaling the label space of object detectors becomes infeasible with a standard supervised approach. This sparked work on the zero-shot setting~\cite{akata_13_attribute_zeroshot,akata_15_embeddings_zeroshot,lampert_09_zeroshot,lampert_14_pami_zeroshot}, where a set of base categories is available at training time, but novel (or unseen) categories are added at test time. While Bansal~\etal~\cite{bansal2018zero} introduced the first work on zero-shot detection, later works relaxed the setting to open-vocabulary~\cite{zareian_cvpr_21}, where annotations other than bounding boxes can be leveraged that may include the novel categories, \eg, image captions~\cite{changpinyo2021conceptual,chen_2015_cococaptions,sharma2018conceptual}. The recent success of large V\&L models~\cite{jia_icml_21_align,khan_eccv22_simla,radford_icml_2021_CLIP} surged interest in open-vocabulary detection~\cite{cai_2022_xdetr,ghiasi_eccv22_openseg,li_cvpr22_glip,gu_iclr_22,lin_23_vldet,zhang_22_glipv2,zhao_eccv22_vlplm}. However, benchmarks for this setting are lacking. Most existing work evaluates on standard detection datasets, COCO~\cite{lin_eccv_14_mscoco} and LVIS~\cite{gupta_cvpr19_lvis}, by separating categories into base and novel. Most recently, \cite{li_22_elevater_odinw} introduces the ODinW benchmark which combines 35 standard detection datasets to setup an open-vocabulary challenge. Still, all benchmarks use a rather limited set of simple category names. In contrast, OmniLabel provides higher complexity with object descriptions being free-form text and, with this, a larger number of unique words (and nouns) which poses a naturally open-vocabulary setting since every description is effectively unique.

%% file: sections/benchmark.tex
\begin{figure*}[t]\centering
    \includegraphics[width=1.0\textwidth]{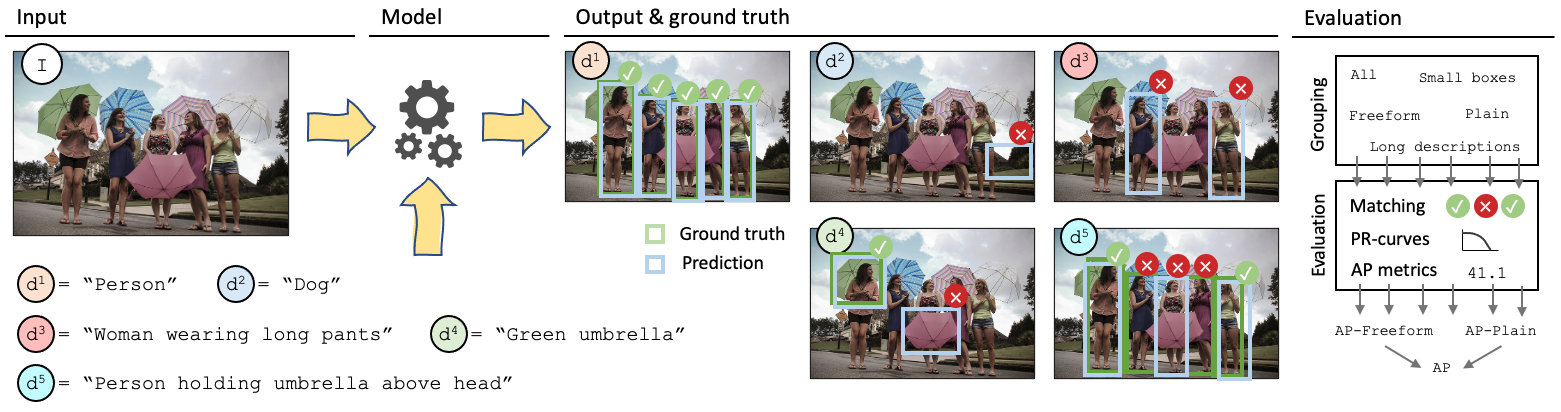}
    \caption{Illustration of the proposed task and evaluation. Given an image $I$ and a set of object descriptions $\left[d^1, d^2, \ldots, d^5\right]$, the model outputs a set of predictions. Each prediction $p^l$ is a triplet, consisting of a bounding box $b^l$ (blue boxes), a confidence score $s^l$ (not shown) and an index $g^l$, which links the prediction to one of the object descriptions $d^*$ (the figure groups predicted boxes by $g^l$). For evaluation, predictions and ground truth (green boxes) are matched based on location (intersection-over-union) as well as the indices $g^l$. In contrast to standard detection, note that one object can be matched with multiple (but different!) descriptions. For example, the woman holding the green umbrella is matched with both $d^1$ and $d^5$. While there is no category-wise grouping as in standard detection, average precision (AP) metrics are computed for other groups, like ``plain categories'', ``free-form object descriptions'', and others. The final metric is the harmonic mean between AP values for ``plain categories'' and ``free-form object descriptions''.}
    \label{fig:overview_task_and_evaluation}
\end{figure*}

\section{Benchmark and Evaluation Metric}
\label{sec:benchmark}
This section provides a formal definition of the benchmark task and the corresponding evaluation metric. An illustration of both is given in \cref{fig:overview_task_and_evaluation}.

\subsection{Benchmark Task}
\label{subsec:task}
\paragraph{Input:} Given a regular RGB image $I_i$ along with a label space $D_i$, the task for model $M$ is to output object predictions $P_i$ according to the label space $D_i$. The subscript in $D_i$ indicates that both content and size $|D_i|$ vary for each image $I_i$. The label space $D_i = \left[ d_i^k \right]_{k=1}^{|D_i|}$ consists of $|D_i|$ elements, $d_i^k \in D_i$, each of which is called an ``object description''. Our object descriptions comprise a combination of plain category names (as in detection) as well as newly-collected free-form text descriptions of objects, see \cref{sec:data_collection}. Being free-form text effectively makes each description unique. While we could define a common label space as the union of all descriptions, this results in a huge label space and poses hard computational challenges on models that tightly fuse image and text, like MDETR~\cite{kamath2021mdetr}. Instead, we vary the label space and each $D_i$ contains both positive (referring to an object in the image) and negative (related to image content but no related objects) descriptions. Examples of free-form object descriptions are given in \cref{fig:omnilabel_examples}.

\paragraph{Output:} Model $M$ must output a set of triplets $P_i = \left[ p_i^l \right]_{l=1}^{|P_i|}$ for image $I_i$ and label space $D_i$. Each triplet $p_i^l = (b_i^l, s_i^l, g_i^l)$ consists of a bounding box $b$, a confidence score $s$, and an index $g$ linking the prediction to an object description in $D_i$. A bounding box $b_i^l$ consists of 4 coordinates in the image space to define the extent of an object, as in standard object detection. The confidence of a model's prediction is expressed by the real-valued scalar $s_i^l$. Finally, the index $g_i^l$ is in the range of $\left[ 1, |D_i| \right]$ and indicates that the prediction $p_i^l$ localizes one object instance of the description $g_i^l$ of the label space $D_i$. Note that multiple predictions $p_i^l$ can point to the same object description $d_i^k$.

\paragraph{Difference to object detection benchmarks:} The main difference is the label space, which is more complex (with natural text object descriptions, often unseen during training) as well as dynamic (size of label space changes for every test image). Standard object detectors fail this task because of their fixed label space assumption.

\paragraph{Difference to referring expression benchmarks:} While the task definition is similar, the key difference is in the corresponding data. First, object descriptions $D_i$ in our benchmark range from plain categories (like in standard detection) to highly specific descriptions. Second, each description can refer to zero, one, or multiple instances in the image. All referring expression datasets assume the presence of the object described by the text and, hence, do not contain negative examples that refer to zero instances, an important aspect of standard detection evaluation. Moreover, only one referring expression dataset (\cite{wu2020phrasecut}) refers to more than a single instance per image.

\subsection{Evaluation Metric}
\label{subsec:evaluation_metric}

To evaluate a model $M$ on our task, we propose a modified version of the object detection evaluation metric, average precision (AP) \cite{lin_eccv_14_mscoco}. This modification is necessary to account for our novel object descriptions that make the label space virtually infinite in size, and that are different for each image. The following list summarizes the changes:
\begin{itemize}
    \item While AP is computed for each category separately (and then averaged) in standard detection, this initial grouping is omitted in OmniLabel. Due to the high specificity of the object descriptions, many of these ``groups'' would then consist of only a single object instance in the whole dataset. This can make the metric less robust. However, to ensure that our metric considers the predicted semantic categories, we adjust the matching between prediction and ground truth. While in standard detection the matching is based purely on the bounding boxes via intersection-over-union (since categories are already grouped), we include the index $g_i^l$ that links a prediction with the object descriptions in $D_i$, see above in \cref{subsec:task}. Specifically, a prediction is matched to a ground truth only if the prediction and the ground truth point to the same object description (semantics) and the predicted and ground truth bounding boxes overlap sufficiently (localization).
    \item Standard detection ground truth exclusively assigns each object instance one semantic category. In contrast, our task requires multi-label predictions. For instance, ``person'' and ``woman in red shirt'' can refer to the same object. This needs to be considered in the matching process of the evaluation. In contrast to standard detection, one ground truth box can be correctly matched with multiple predictions if the match happens via different object descriptions (recall index $g_i^l$ above).
    \item Our object descriptions $D_i$ contain both plain category names (like ``car'' or ``person'' from standard detection) as well as complex free-form text (like ``blue sports car parked near left sidewalk''). We want our metric to give equal importance to both types. Due to the different number of ground truth instances, we first compute AP for both types separately and then take the harmonic mean. Different from the arithmetic mean, the harmonic mean requires good results on both types to achieve a high number on the final metric.
\end{itemize}
%
We implemented this evaluation protocol in Python and released it at \url{https://github.com/samschulter/omnilabeltools}

%% file: sections/datacollection.tex
\section{Dataset Collection}
\label{sec:data_collection}

\begin{figure}[t]\centering
    \includegraphics[width=1.0\columnwidth]{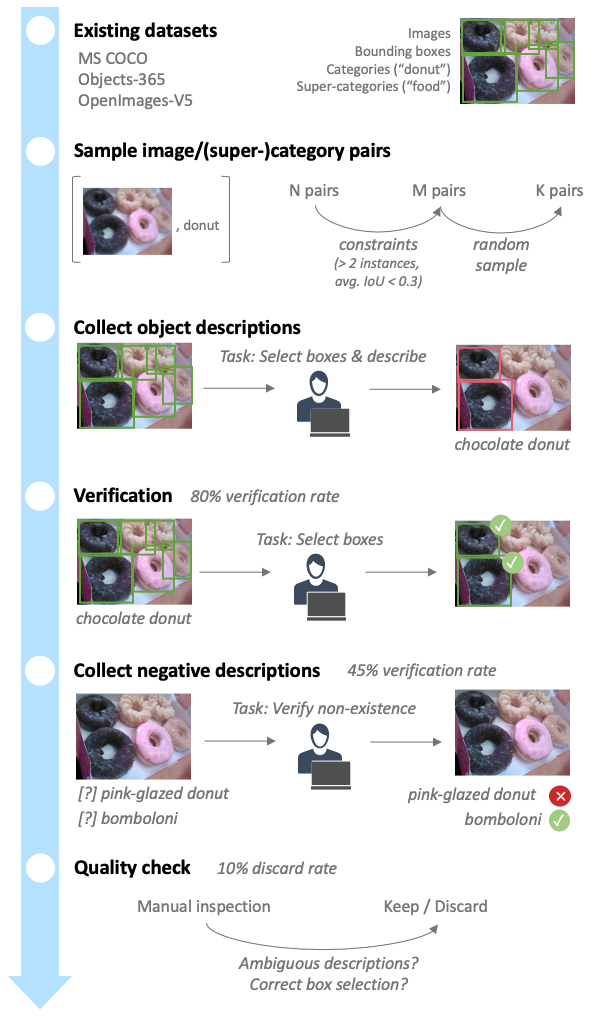}
    \caption{A step-by-step summary of our data collection and annotation process. \cref{sec:data_collection} describes each step in detail. The key benefits of our annotation process are: (1) reuse of existing object detection datasets with annotated categories allows for a balanced sampling and consequently a set of diverse object description. (2) Contrary to existing free-form text benchmarks, we collect negative descriptions, which are related to an image but do not refer to any object.}
    \label{fig:data_collection}
\end{figure}

To establish our novel evaluation benchmark, we need images that annotate objects with bounding boxes and corresponding free-form text descriptions. To do so, we define a multi-step annotation and verification process. \cref{fig:data_collection} and the following paragraphs describe the process.

\paragraph{Existing datasets:} We start with the validation/test sets of COCO~\cite{lin_eccv_14_mscoco}, Objects-365~\cite{Objects365}, and OpenImages-V5~\cite{OpenImages}, which not only saves annotation cost for obtaining bounding boxes, but also helps collecting \emph{diverse} object descriptions. By leveraging the (super-)category information when sampling images, we force annotators to provide descriptions also for rare categories. Otherwise, annotators will quickly pick simple and common categories to describe. And although reusing datasets may exclude some categories that were not annotated, the object descriptions we collect often include additional categories. For example, while a ``bottle cap'' is not part of the original categories, a description like ``bottle with red cap'' requires to understand ``bottle cap''.

\paragraph{Sample image / (super)category pairs:} 
To encourage a diverse distribution of categories and images, we propose a strategy to randomly sample pairs of images and (super) categories. We first filter all possible pairs based on the following criteria: (a) At least two instances of a (super) category need to be present in the image. (b) For super-categories, at least two different sub-categories need to be present in the image. (c) To collect descriptions that focus on the object's appearance, relations and actions, we reject pairs with more than 10 instances, if the larger side of any instances' bounding box is smaller than 80 pixels, if the average over the bounding box's largest overlap with any other box is larger than 50\%, or if any instance is flagged as covering a crowd of objects (``iscrowd''). Finally, we pick a random subset of the filtered pairs for annotation with free-form descriptions.

\paragraph{Collect object descriptions:} All initial object descriptions are collected with Amazon Mechanical Turk (AMT). Given an image/(super-)category pair, we draw the bounding boxes of that (super-)category's instances and request annotators to pick a subset of the instances and provide a text description that only matches their selection. Specifically, for {\emph image/category} pairs with $N=2$ possible instances, we ask to pick exactly one. If $N>2$, we ask to select at least 2 but at most $N-1$ instances. This ensures that if an object description uses the category name itself, additional text is needed to distinguish the instances. For {\emph image/super-category} pairs, we ask to select at least one instance, but avoid using the category names themselves in the descriptions. This encourages higher-level descriptions like ``edible item'' for all objects of the super-category ``food''. Finally, we ran a semi-automatic profanity check~\cite{profanity_check} on the collected text. We manually inspected 500 descriptions with the highest probability of containing profane language, but did not need to discard any description.

\paragraph{Verification:} To ensure high quality descriptions, we again use AMT to verify the selection of bounding boxes from the previous step. We provide annotators the originally highlighted bounding boxes and the newly collected description and request to select the objects for which the description applies. We only keep descriptions for which both selections (initial and verification) are equal, which is about 80\% of all descriptions.

\paragraph{Collect negative descriptions:} As described earlier, a key aspect of our benchmark are negative object descriptions. These descriptions are related to an image, but do not actually refer to any object. To collect such descriptions, we leverage the already-collected free-form object descriptions with their underlying (super-)category information. Hence, a sample \& verify approach is suitable, where, for each image/(super-)category pair, we randomly sample 5 object descriptions from the same (super-)category but a different image and ask 2 AMT annotators to confirm that the given description does not refer to any object. We then only keep negative descriptions with 2 confirmations, which was about 30\% in our case.

\paragraph{Quality check:} Finally, we perform a manual quality check. We fix misspellings and ambiguous descriptions when possible. If the meaning of the description changed, we keep the positive description, but discard all negative associations to other images. If an object description is entirely wrong, we discard it, which was the case for about 10\% of the remaining descriptions.

\paragraph{Annotators:} In total, 263 different annotators from AMT provided inputs for our annotations. For the three tasks using AMT (generating descriptions, verifying descriptions, and verifying negative descriptions), we had 54, 71, and 235 annotators, respectively. 

%% file: sections/dataanalysis.tex
\begingroup
\setlength{\tabcolsep}{3.5pt} 
\renewcommand{\arraystretch}{1.15} 
\begin{table}[t]\centering\footnotesize
\begin{tabular}{lllll}
  & \rotatebox{0}{RefCOCO/g/+} & \rotatebox{0}{Flickr30k} & \rotatebox{0}{PhraseCut} & \rotatebox{0}{\textbf{OmniLabel}}   \\
\toprule
\# images       & 4.3K        & 1.0K         & 2.9K        & 12.2K   \\
\# descr.       & 26.5K       & 11.3K        & 19.5K       & 15.8K {\tiny (16.8K)}   \\
$\;\;$ \# pos   & 26.5K       & 11.3K        & 19.5K       & 11.7K      \\
$\;\;$ \# neg   & 0           & 0            & 0           & 9.4K         \\
\# boxes        & 10.2K       & 4.6K         & 32.1K       & 20.4K {\tiny (165.7K)}  \\
\# boxes/descr  & 1.0$\pm$0.0 & 1.0$\pm$0.0  & 1.6$\pm$1.6 & 1.7$\pm$1.0      \\
\bottomrule
\end{tabular}
    \caption{
    Basic statistics of the validation sets of OmniLabel, the combination of RefCOCO/g/+~\cite{mao_cvpr16_refcocog,yu_eccv16_refcoco}, Flickr30k~\cite{flickrentitiesijcv} and PhraseCut~\cite{wu2020phrasecut}.
    All numbers are based only on free-form object descriptions. Numbers in parenthesis in the last column indicate statistics with plain categories included
    }
    \label{tab:omnilabel_basic_stats}
\end{table}
\endgroup

\section{Dataset Analysis}
\label{sec:analysis}
This section analyzes various statistics of OmniLabel and compares them with other related datasets. For all datasets, we analyze the corresponding validation sets.

\subsection{Basic statistics}
\label{subsec:analysis_basic_stats}
\cref{tab:omnilabel_basic_stats} summarizes key numbers of our OmniLabel dataset in comparison with prior benchmarks on referring expressions or visual grounding, specifically, the combination of RefCOCO/g/+~\cite{mao_cvpr16_refcocog,yu_eccv16_refcoco}, Flickr30k~\cite{flickrentitiesijcv} and PhraseCut~\cite{wu2020phrasecut}.
The key takeaways are: (a) The existence of negative objet descriptions (\# neg). Like in standard detection, where categories not present in an image are considered negative and are still evaluated, OmniLabel provides free-form object descriptions that are related to the image but do not refer to any object. (b) The number of bounding boxes per description is higher than for any other dataset, which adds to the difficulty of the benchmark.

\subsection{Analysis of free-form object descriptions}
\label{subsec:analysis_freeform_descriptions}
While OmniLabel also contains plain categories like in standard object detection, our focus for this analysis is on the free-form object descriptions.

\begin{figure}
    \centering
    \includegraphics[width=1.0\columnwidth]{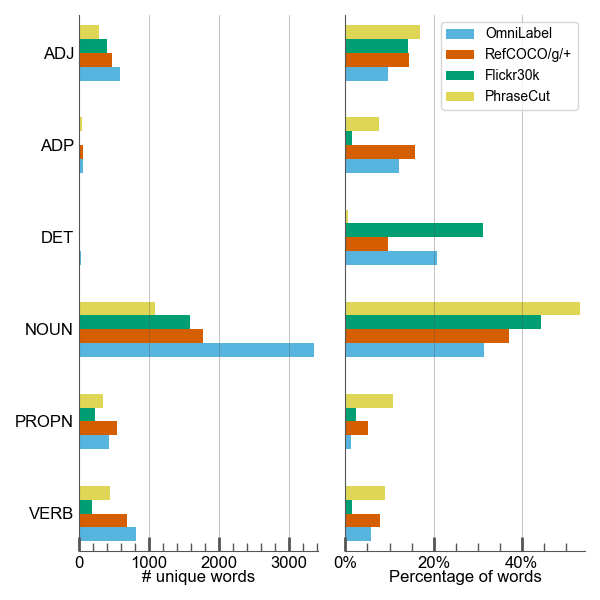}
    \caption{
    Grouping of the words in our object descriptions into relevant part-of-speech tags. We show the count of unique words (left) and the distribution of words (right). The statistics are computed from a random subset of 10K descriptions of each dataset.
    }
    \label{fig:pos_tag_histograms}
\end{figure}

\paragraph{Part-Of-Speech (POS) tagging:}
To analyze the content and the diversity of our object descriptions, \cref{fig:pos_tag_histograms} shows an analysis of the words when grouped by part-of-speech tagging. On the left, we have the number of unique words (not counting multiple occurrences) based on a random subset of 10K descriptions. For adjectives, verbs and particularly nouns, OmniLabel covers more unique words, attributing to its diversity. On the right, we have the distribution of (non-unique) words among the different POS tags. We observe a more uniform distribution than other datasets, indicating longer descriptions (see below) that are closer to sentences, rather than short phrases or single words. On average, we have 2.04 $\pm$ 0.90 nouns, 0.62 $\pm$ 0.71 adjectives and 0.43 $\pm$ 0.61 verbs per object description.

\paragraph{Description lengths:} \cref{fig:histogram_description_lengths} confirms our assumption from above that object descriptions in OmniLabel contain more words than other datasets.


%% file: sections/baselines.tex
\section{Baselines}
\label{sec:baselines}
Beyond statistics of the collected annotations, we also evaluate recent language-based object detection models on OmniLabel with our novel evaluation metric.

\subsection{Models}
\label{subsec:baseline_models}
Our evaluation aims to encompass a wide range of models, and we select them based on their performance on (a) standard detection benchmarks (like LVIS~\cite{gupta_cvpr19_lvis} and COCO~\cite{lin_eccv_14_mscoco}), and (b) tasks like Phrase Grounding and Referring Expression Compression. For models that primarily focus on open-vocabulary detection via large-scale pre-training, we utilize RegionCLIP~\cite{zhong2022regionclip} and Detic~\cite{zhou2021detic}. For models that are designed for text-conditioned detection with state-of-the-art performance on visual grounding, we use MDETR~\cite{kamath2021mdetr}, GLIP~\cite{li_cvpr22_glip}, and FIBER~\cite{dou2022coarse}.
We present a brief summary of each of these models.

\begin{figure}[t!]
    \centering
    \includegraphics[width=1.0\columnwidth]{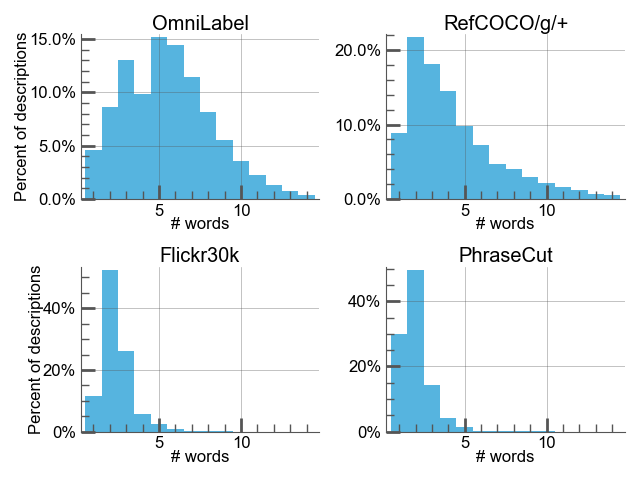}
    \caption{Histograms of description lengths (in number of words) for OmniLabel and three other datasets.}
    \label{fig:histogram_description_lengths}
\end{figure}

\noindent
{\bf RegionCLIP} is an open-vocabulary object detector based on Faster RCNN \cite{Ren_2015_NIPS}. It adopts pretrained CLIP's visual encoder (ResNet-50) \cite{radford_icml_2021_CLIP} as the backbone and is finetuned with image-text pairs from the Internet (e.g. CC3M \cite{sharma2018conceptual}). Thus, RegionCLIP is expected to get lower performance on our benchmark, compared to other baselines trained with detection and visual grounding datasets.

\noindent
\textbf{Detic} is an open-vocabulary object detector that relies on CLIP \cite{radford_icml_2021_CLIP} embeddings to encode class names. It utilizes a combination of box-level and image-level annotations, with a loss function that is weakly-supervised (modified Federated Loss \cite{zhou2021probabilistic}). For the results presented, we utilized Swin-Base \cite{liu2021swin} as the backbone.

\noindent
{\bf MDETR} is an end-to-end modulated detector that can detect objects for a given free-form text query, with a tight coupling between image and text modalities. The model is based on DETR \cite{carion2020end} and trained with a combination of different visual grounding datasets (GoldG).

\noindent
{\bf GLIP} is a large scale visual grounding model that is trained with a combination of detection annotations, visual grounding data and image-text pairs. We evaluate its two variants, GLIP-T and GLIP-L. GLIP-T adopts Swin-Tiny \cite{liu2021swin} as the backbone and is trained with Objects365 \cite{Objects365}, GoldG \cite{kamath2021mdetr}, CC3M, and SBU \cite{ordonez2011im2text}. GLIP-L adopts Swin-Large \cite{liu2021swin} as the backbone and is trained with several detection datasets (including Objects365, OpenImages \cite{OpenImages}, and Visual Genome \cite{krishna2017visual}), GoldG, CC12M \cite{changpinyo2021conceptual}, SBU, and additional 24M image-text pairs collected from the Internet.

\noindent
{\bf FIBER-B} introduces a two-stage pretraining strategy, from coarse- to fine-grained data, with image-text and image-text-box annotations, respectively. It generally follows the model design and the training protocol of GLIP, but adopts Swin-Base \cite{liu2021swin} as the backbone.

\begingroup
\setlength{\tabcolsep}{4pt} 
\renewcommand{\arraystretch}{1.15} 
\begin{table}\centering
\footnotesize
\begin{tabular}{clccccccc}
\rotatebox{90}{Images} & \rotatebox{90}{Method} & \rotatebox{90}{AP} & \rotatebox{90}{AP-categ} & \rotatebox{90}{AP-descr} & \rotatebox{90}{AP-descr-pos} & \rotatebox{90}{AP-descr-S} & \rotatebox{90}{AP-descr-M} & \rotatebox{90}{AP-descr-L} \\
\toprule
\multirow{6}{*}{\rotatebox{90}{All}}
    & RegionCLIP~\cite{zhong2022regionclip} & 2.7 & 2.7 & 2.6 & 3.2 & 3.6 & 2.7 & 2.3 \\  
    & Detic~\cite{zhou2021detic} & 8.0 & 15.6 & 5.4 & 8.0 & 5.7 & 5.4 & 6.2 \\  
    \cline{2-9}
    & MDETR~\cite{kamath2021mdetr} & - & - & 4.7 & 9.1 & 6.4 & 4.6 & 4.0 \\  
    & GLIP-T~\cite{li_cvpr22_glip} & 19.3 & 23.6 & 16.4 & 25.8 & 29.4 & 14.8 & 8.2 \\  
    & GLIP-L~\cite{li_cvpr22_glip} & 25.8 & 32.9 & 21.2 & 33.2 & 37.7 & 18.9 & 10.8 \\ 
    & FIBER-B~\cite{dou2022coarse} & 25.7 & 30.3 & 22.3 & 34.8 & 38.6 & 19.5 & 12.4 \\ 
    \midrule    
\multirow{6}{*}{\rotatebox{90}{COCO}}
    & RegionCLIP & 4.1 & 5.1 & 3.5 & 5.1 & 6.1 & 3.3 & 4.1  \\  
    & Detic & 8.3 & 43.1 & 4.6 & 9.9 & 10.2 & 3.5 & 7.2 \\  
    \cline{2-9}
    & MDETR & - & - & 13.2 & 31.6 & 15.4 & 13.5 & 12.4 \\  
    & GLIP-T & 18.7 & 45.7 & 11.7 & 31.2 & 27.0 & 10.9 & 10.2 \\  
    & GLIP-L & 21.8 & 50.4 & 13.9 & 36.8 & 28.9 & 12.9 & 11.5 \\ 
    & FIBER-B & 22.2 & 49.6 & 14.3 & 38.8 & 31.3 & 12.7 & 14.2 \\ 
    \midrule    
\multirow{6}{*}{\rotatebox{90}{Objects-365}}
    & RegionCLIP & 3.6 & 3.6 & 3.6 & 4.1 & 5.0 & 3.5 & 3.0 \\  
    & Detic & 9.1 & 21.6 & 5.7 & 8.4 & 6.6 & 5.9 & 6.9 \\  
    \cline{2-9}
    & MDETR & - & - & 3.2 & 5.9 & 3.0 & 3.2 & 2.7 \\  
    & GLIP-T & 22.6 & 30.0 & 18.1 & 26.9 & 34.2 & 16.0 & 9.1 \\  
    & GLIP-L & 29.3 & 37.5 & 24.0 & 35.2 & 44.5 & 20.5 & 11.8 \\ 
    & FIBER-B & 30.8 & 37.9 & 25.9 & 38.2 & 44.7 & 22.5 & 14.1 \\ 
    \midrule    
\multirow{6}{*}{\rotatebox{90}{OpenImages v5}}
    & RegionCLIP & 2.3 & 2.1 & 2.7 & 2.9 & 3.4 & 2.7 & 2.0 \\  
    & Detic & 6.4 & 8.1 & 5.4 & 6.9 & 5.4 & 5.6 & 5.8 \\  
    \cline{2-9}
    & MDETR & - & - & 6.1 & 10.6 & 9.6 & 5.7 & 4.1 \\  
    & GLIP-T & 17.6 & 20.0 & 15.7 & 24.4 & 25.8 & 14.9 & 7.5 \\  
    & GLIP-L & 25.7 & 35.8 & 20.1 & 31.2 & 33.3 & 18.7 & 10.3 \\ 
    & FIBER-B & 22.0 & 24.4 & 20.1 & 30.9 & 34.1 & 18.5 & 10.5 \\ 
    \bottomrule
\end{tabular}
\caption{
Evaluation of language-based detection baselines on the OmniLabel benchmark with the metric described in \cref{subsec:evaluation_metric}. The final AP value is the geometric mean of only plain categories (AP-categ) and free-form descriptions (AP-descr). AP-descr-pos evaluates on only positive descriptions, clearly showcasing the impact of negative descriptions. AP-descr-S/M/L evaluate descriptions of different length (up to 3 words, 4 to 8, and more than 8)
}
\label{table:omni_baseline2}
\end{table}
\endgroup

\subsection{Results}
\label{subsec:baseline_results}
We run two experiments with the above described models. The first one focuses on a detailed analysis of our new metric (see \cref{subsec:evaluation_metric}) on the OmniLabel dataset. The second experiment compares our metric on three different datasets.

\paragraph{Analysis on the OmniLabel dataset -- \cref{table:omni_baseline2}:}
The first observation we make is that nearly all methods achieve higher accuracy on plain object categories (AP-categ) compared to free-form text descriptions (AP-descr).
One can also clearly see the effect of using the geometric mean for the final metric (AP), when averaging over plain categories (AP-categ) and free-form descriptions (AP-descr). This effect is more pronounced for COCO images.

A key takeaway message from \cref{table:omni_baseline2} is the impact of negative descriptions. The performance gap between including negative descriptions in the label space (AP-descr) and excluding them (AP-descr-pos) is significant. The biggest gap can be observed for COCO images, which is because this part of the dataset contains the most negative descriptions relative to the number of images (due to our data collection process),
see \cref{sec:supp_data_collection}).
Another observation we get from \cref{table:omni_baseline2} is that accuracy correlates negatively with description length. AP values are in general higher for shorter descriptions (AP-descr-S, up to three words) than for longer descriptions (AP-descr-L, more than 8 words).

Finally, we can see that GLIP-T/L and FIBER-B achieve the best results on OmniLabel. MDETR achieves reasonable results when only considering positive descriptions (AP-descr-pos) but fails when negatives are added (AP-descr), likely due to the specific training algorithm. Also, we did not report results of MDETR for AP or AP-categ due to the significant runtime induced by the large labelspace and MDETR's model design. As expected, Detic is good on plain categories (AP-categ) but underperforms on free-form descriptions (AP-descr). RegionCLIP's lower performance is likely due to a combination of a weaker backbone and the training data.

\begingroup
\setlength{\tabcolsep}{5pt} 
\renewcommand{\arraystretch}{1.15} 
\begin{table}[ht]\centering
\small
\begin{tabular}{lcccc}
Method & \multicolumn{2}{c}{OmniLabel} & RefCOCOg & PhraseCut  \\
       & {\footnotesize descr} & \footnotesize{descr-pos} & {\footnotesize descr} & {\footnotesize descr}  \\
\toprule
        RegionCLIP~\cite{zhong2022regionclip} & 2.6 & 3.2 & 1.1 & 2.2 \\
        Detic~\cite{zhou2021detic} & 5.4 & 8.0 & 6.8 & 6.8 \\
        \midrule 
        GLIP-T~\cite{li_cvpr22_glip} & 16.4 & 25.8 & 32.1 & 23.9 \\
        GLIP-L~\cite{li_cvpr22_glip} & 21.2 & 33.2 & 33.4 & 29.3 \\
        FIBER-B~\cite{dou2022coarse} & 22.3 & 34.8 & 33.0 & 27.4 \\
    \bottomrule
\end{tabular}
\caption{
Comparing our evaluation metric (\cref{subsec:evaluation_metric}) for all models on three different datasets. OmniLabel poses a more difficult challenge, specifically when negative descriptions are included (descr vs. descr-pos)
}
\label{table:omni_difficulty}
\end{table}
\endgroup

\paragraph{Evaluation metric across different datasets -- \cref{table:omni_difficulty}:} 
We compare all models on three datasets (OmniLabel, RefCOCOg~\cite{mao_cvpr16_refcocog} and PhraseCut~\cite{wu2020phrasecut}). We make two main observations: First, OmniLabel is a more difficult benchmark, particularly because of negative descriptions. Second, the proposed evaluation metric from \cref{subsec:evaluation_metric} is more stringent than the one used in RefCOCO/g/+. For instance, FIBER-B on RefCOCOg (val) achieves 87.1\% accuracy \cite{dou2022coarse} compared to the 33.0 from \cref{table:omni_difficulty}.

%% file: sections/conclusion.tex
\section{Conclusions}
\label{sec:conclusion}
OmniLabel presents a novel benchmark for evaluating language-based object detectors. A key innovations is the annotation process, which (a) encourages free-form text descriptions of objects that are complex and diverse, (b) ensures collecting difficult examples with multiple instances of the same underlying category present in the images, and (c) provides negative free-form descriptions that are related but not present in an image. Moreover, OmniLabel defines a novel task setting and a corresponding evaluation metric.
Our analysis of the dataset shows that we could indeed collect object descriptions that are diverse and contain more unique nouns, verbs and adjectives than existing benchmarks. Also, evaluating recent language-based object detectors confirmed the level of difficulty that OmniLabel poses to these models.
We hope that our contributions in providing a challenging benchmark help progress the field towards robust object detectors that understand semantically rich and complex descriptions of objects.

%% file: sections_supp/supp_dataanalysis.tex
\section{Additional dataset analysis}
\label{sec:supp_data_analysis}
We further analyze the object descriptions we collected for the OmniLabel benchmark in the following paragraphs.

\begin{figure}[t]\centering
    \begin{subfigure}[b]{1.0\columnwidth}\centering
    \includegraphics[width=1.0\textwidth]{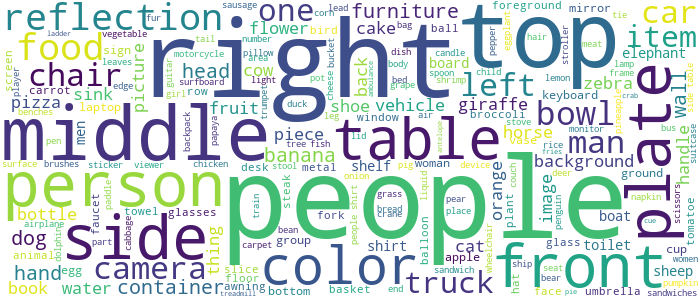}
    \caption{}
    \label{fig:word_cloud_noun}
    \end{subfigure}
    \hfill
    \begin{subfigure}[b]{1.0\columnwidth}\centering
    \includegraphics[width=1.0\textwidth]{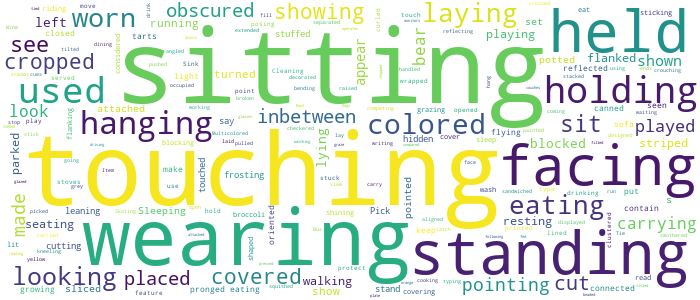}
    \caption{}
    \label{fig:word_cloud_verb}
    \end{subfigure}
    \hfill
    \begin{subfigure}[b]{1.0\columnwidth}\centering
    \includegraphics[width=1.0\textwidth]{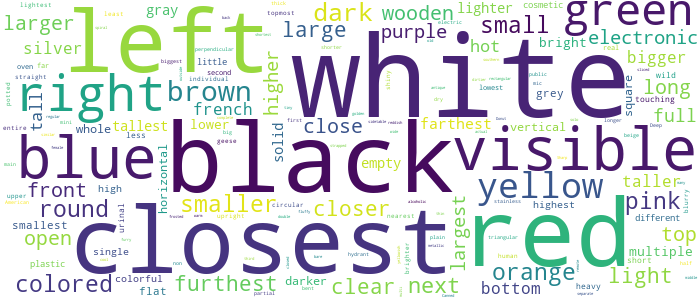}
    \caption{}
    \label{fig:word_cloud_adj}
    \end{subfigure}
    \caption{Word clouds of nouns (a), verbs (b) and adjectives (c) collected from a subset of 5K object descriptions.}
    \label{fig:word_clouds}
\end{figure}

\paragraph{Part-Of-Speech (POS) tags:}
In~\cref{subsec:analysis_freeform_descriptions}, 
we analyze object descriptions by their POS tagging. To get POS tags, we use the \texttt{spacy} toolbox~\cite{spacy_toolbox}, which categorizes each word into one of 17 UPOS tags~\cite{upos_tags}, out of which we selected the 6 most relevant tags for 
\cref{fig:pos_tag_histograms}:
\begin{itemize}
    \item ADJ: adjective
    \item ADP: adposition
    \item DET: determiner
    \item NOUN: noun
    \item PROPN: proper noun
    \item VERB: verb
\end{itemize}
\cref{fig:word_clouds} shows word clouds for the tags NOUN, VERB and ADJ, collected from a random subset of 5K object descriptions.

\begin{figure}[t]\centering
    \includegraphics[width=1.0\columnwidth]{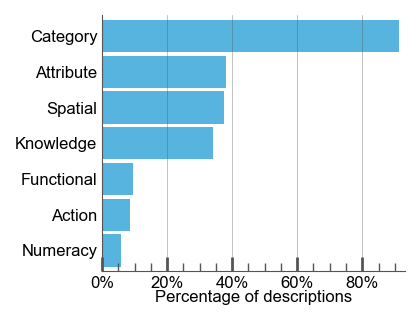}
    \caption{
    We manually tagged a random subset of 500 object descriptions with the types of language understanding needed to localize the referred instances correctly. The plot shows the percentage of object descriptions tagged for each type. Each description can be tagged with multiple types.
    }
    \label{fig:histogram_descr_group_by_manual_types}
\end{figure}

\paragraph{Types of language understanding:}
To further analyze the our object descriptions, we manually tagged a random subset of 500 descriptions with what type of language understanding they require:
\begin{itemize}
    \item \textbf{``categories''}: The description contains one or more object category names
    \item \textbf{``spatial relations''}: Example: ``left to'', ``behind''
    \item \textbf{``attributes''}: Attribute of objects, \eg, color or material
    \item \textbf{``(external) knowledge or reasoning''}: Knowledge beyond the image content
    \item \textbf{``functional relations''}: Describing objects by their functionality, \eg: ``edible item'' or ``areas to sit on''
    \item \textbf{``actions''}: Any action an object can perform, ``person jumping'', ``parked car''
    \item \textbf{``numeracy''}: Descriptions that require reasoning about numbers, like counting or understanding the time
\end{itemize}
\cref{fig:histogram_descr_group_by_manual_types} shows the results of our manual tagging efforts as the percentage of description that were tagged with one of the above types. Note that one description can be tagged with multiple categories. For example, the description ``A black cat jumping onto the chair on the left'' would get tags for ``attribute'' (black), ``categories'' (cat, chair), ``action'' (jumping), and ``spatial relations'' (on the left).

We can see from \cref{fig:histogram_descr_group_by_manual_types} that more than 80\% of object description include some category name, which is expected. Note that the number of unique nouns is not limited to a fixed label space. In fact, the validation set of OmniLabel has 4.6K unique nouns, a lot more than existing benchmarks,
see~\cref{tab:dataset_stats_basic}.
Besides category names, close to 40\% of object descriptions require an understanding of attributes, spatial relations, and external knowledge or reasoning for correct localization of objects. And finally, understanding of functional capabilities, actions and numeracy is needed for 5-10\% of the descriptions. In \cref{sec:supp_dataset_examples}, we provide visual examples for each of the above groups.

\begin{figure}\centering
    \includegraphics[width=1.0\columnwidth]{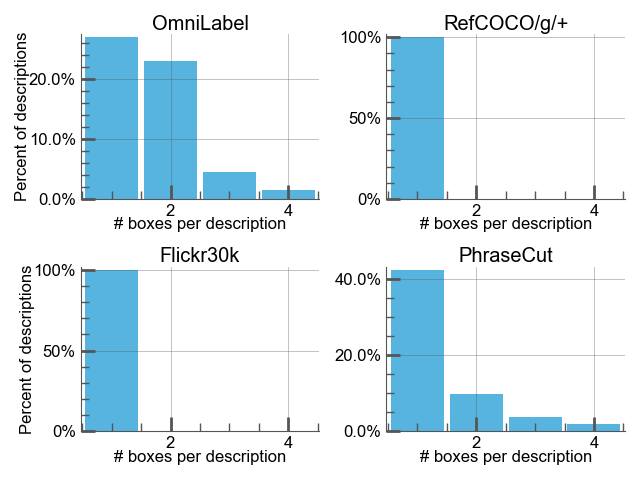}
    \caption{
    Distribution of object descriptions referring to different number of instances in the image.
    }
    \label{fig:hist_num_boxes_per_description}
\end{figure}

\paragraph{Distribution of number of boxes per description:}
One aspect that differentiates our OmniLabel dataset from prior benchmarks is the number of instances (bounding boxes) that are referred to by one object description. As we can see in \cref{fig:hist_num_boxes_per_description}, for both  RefCOCO/g/+~\cite{mao_cvpr16_refcocog,yu_eccv16_refcoco} and Flickr30k~\cite{flickrentitiesijcv} all descriptions refer to exactly one instance in the image. PhraseCut~\cite{wu2020phrasecut} and OmniLabel allow multiple instances per description, while OmniLabel shows a lower bias towards referring to one instance.

%% file: sections_supp/supp_datacollection.tex
\section{Additional information on data collection}
\label{sec:supp_data_collection}
\cref{sec:data_collection}
describes our data collection process. One aspect of this process is that we start from object detection datasets with existing annotations of bounding boxes and corresponding semantic categories.
On COCO~\cite{lin_eccv_14_mscoco}, semantic annotations contain a category name along with a grouping into super-categories. For Objects-365~\cite{Objects365} and OpenImages~\cite{OpenImages}, we manually grouped categories into super-categories.

\begin{figure}[t]\centering
    \includegraphics[width=1.0\columnwidth]{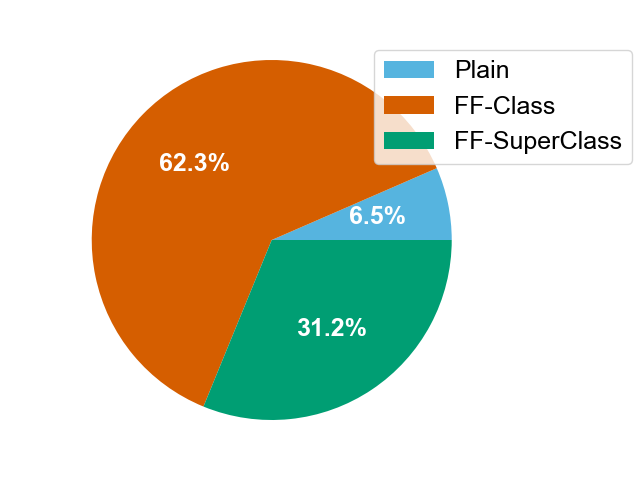}
    \caption{
    Pie chart showing the distribution of object descriptions grouped into plain categories and free-form descriptions collected based on standard categories (FF-Class) and super-categories (FF-SuperClass).
    }
    \label{fig:piechart_descr_group_by_annotype}
\end{figure}

We leverage this semantic annotation when selecting images for annotation with free-form object descriptions. Specifically, we sample pairs of images and (super-) categories for annotation that fulfill some constraints
(enough instances available, see \cref{sec:data_collection}).
\cref{fig:piechart_descr_group_by_annotype} shows the distribution of object descriptions over their origin:
\begin{itemize}
    \item Plain: Original categories of the underlying dataset
    \item FF-Class: Free-form descriptions based on categories
    \item FF-SuperClass: Free-form descriptions based on super-categories
\end{itemize}
The intuition behind sampling based on different types of categories is to collect object descriptions that go beyond using the original category names along with additional context to specify subsets of object instances. And indeed, we found that 45.3\% of the ``FF-Class'' descriptions use the underlying category name, while only 10.8\% of the ``FF-SuperClass'' descriptions use the super-category name and only 5.3\% of the ``FF-SuperClass'' descriptions use any of the subclass names.

\begin{figure}[t]\centering
    \includegraphics[width=1.0\columnwidth]{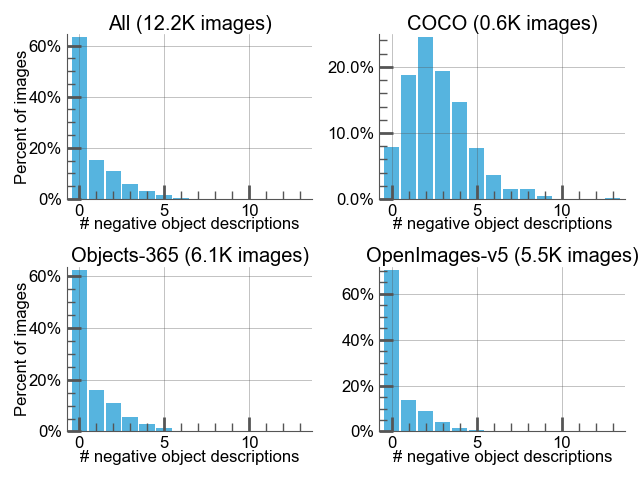}
    \caption{
    Distribution of images with different number of negative descriptions. The title of each sub-plot indicates the subset of images that were inspected. Images from COCO have a significantly different distribution to Objects-365 and OpenImages-v5 due to our annotation schedule, see text.
    }
    \label{fig:histogram_negative_descriptions_per_image}
\end{figure}

\paragraph{Collection of negative object descriptions}
A major claim in our paper is the existence of negative descriptions. These are object descriptions that are semantically related to an image, but do not refer to any object. For any given image, we collect such negative descriptions by first randomly sampling collected positive descriptions from other images that contain the same (super-) category. Then, the randomly selected descriptions are manually verified by human annotators to not refer to any object in the image. The semantic relation to the image we obtain from the sampling process makes these negative descriptions difficult distractors. \cref{fig:ex_neg_a,fig:ex_neg_b} in \cref{sec:supp_dataset_examples} show several examples.

Finally, \cref{fig:histogram_negative_descriptions_per_image} shows a distribution of the number of negatives per image, for all images of the dataset as well as for the set of images coming from the three datasets we used for annotation, COCO~\cite{lin_eccv_14_mscoco}, Objects-365~\cite{Objects365}, and OpenImages-v5~\cite{OpenImages}. The figure shows a significantly different distribution for COCO compared to the other datasets. The absolute numbers of negatives are different given the number of images per dataset, see title of sub-plots. Still, there are two reasons for this stark difference and both relate to our annotation process. First, we collected negative descriptions only for 50\% of the images in Objects-365 and OpenImages-v5\footnote{This might change in the future when we collect more data}. Second, we found that the verification rate of negative descriptions
(see \cref{sec:data_collection})
is clearly higher for COCO (around 45\%) compared to Objects-365 (around 25\%) and OpenImages-v5 (around 16\%). We suspect the number of underlying object categories to cause this difference in the verification rates, but this aspect needs further investigation.

Nevertheless, the total number of negative descriptions in OmniLabel is currently around 10K, sufficient to make a clearly noticeable impact in the evaluation of models. This can be seen from our evaluation
in \cref{table:omni_baseline2},
specifically when looking at the difference between AP-descr and AP-descr-pos. The difference between these metrics is that AP-descr-pos does not evaluate on negative descriptions. Given that we observe significantly higher numbers for AP-descr-pos, particularly for COCO images, we can safely conclude that negative descriptions pose a significant challenge to current language-based models.

\paragraph{Annotation interface:} We provide screenshots of the annotation interface for the three tasks we rely on human annotators,
see also \cref{fig:data_collection}:
\begin{enumerate}[label=(\alph*)]
    \item ``collect object descriptions'' (\cref{fig:amt_collect_descriptions})
    \item ``Verification of descriptions'' (\cref{fig:amt_verify_descriptions})
    \item ``Collection of negative descriptions'' (\cref{fig:amt_verify_negatives}).
\end{enumerate}

%% file: sections_supp/supp_code_and_dataset.tex
\section{Code and Dataset}
\label{sec:supp_code_and_dataset}
\begin{figure}[t]\centering
    \begin{subfigure}[b]{0.46\columnwidth}\centering
    \includegraphics[height=9.6em]{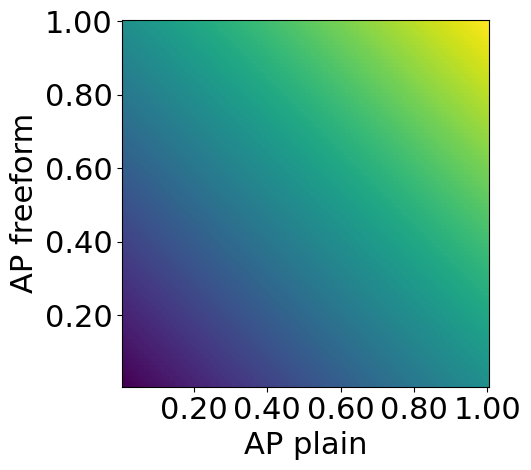}
    \caption{}
    \label{fig:arithmetic}
    \end{subfigure}
    \hfill
    \begin{subfigure}[b]{0.52\columnwidth}\centering
    \includegraphics[height=9.6em]{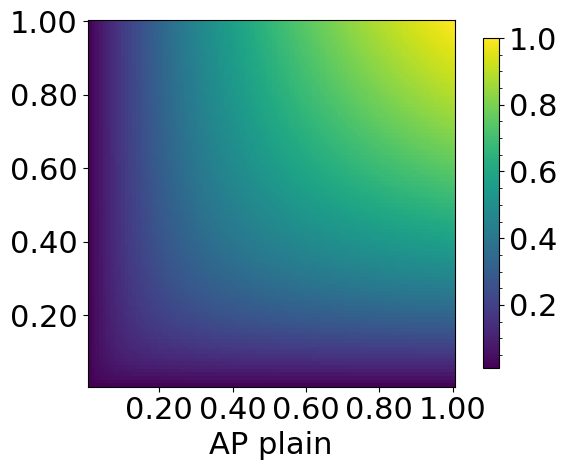}
    \caption{}
    \label{fig:harmonic}
    \end{subfigure}
    \caption{Difference between (a) arithmetic and (b) harmonic mean when averaging two values. The preferred choice in our metric to average AP of plain and freeform-text object descriptions is the harmonic mean, because it encourages good performance on both types of object descriptions. This is apparent from the low values in both the upper left and lower right corners in (b).}
    \label{fig:metric_harmonic_mean}
\end{figure}

Along with the dataset, we built a Python-based toolkit to visualize samples from the dataset and to evaluate prediction results.
The toolbox is publicly released at \url{https://github.com/samschulter/omnilabeltools} and includes a Jupyter notebook \texttt{omnilabel\_demo.ipynb} demonstrating the use of the library. The last cell in the notebook runs the evaluation with dummy predictions. The final metric,
as described in \cref{subsec:evaluation_metric},
is the harmonic mean between AP for plain and freeform-text object descriptions. \cref{fig:metric_harmonic_mean} illustrates the impact of using the harmonic mean over the arithmetic mean.



%% file: sections_supp/supp_dataset_examples.tex
\section{Examples of Dataset Samples}
\label{sec:supp_dataset_examples}
\begin{figure*}[t]\centering
    \begin{subfigure}[b]{0.48\textwidth}\centering
    \includegraphics[width=1.0\textwidth]{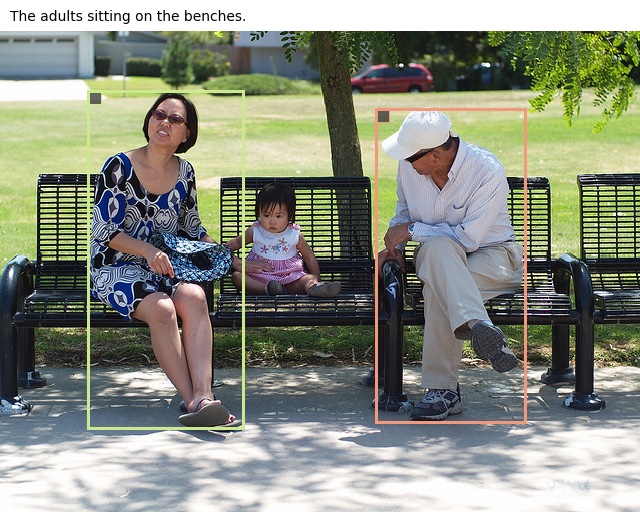}
    \caption{}
    \label{fig:ex_c_01}
    \end{subfigure}
    \hfill
    \begin{subfigure}[b]{0.48\textwidth}\centering
    \includegraphics[width=1.0\textwidth]{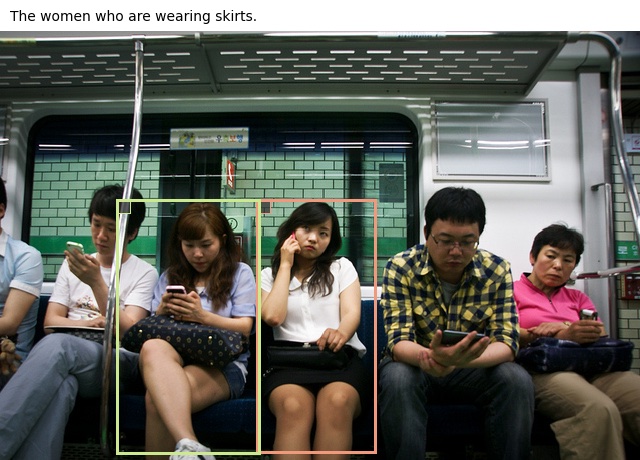}
    \caption{}
    \label{fig:ex_d_02}
    \end{subfigure}
    \\
    \begin{subfigure}[b]{0.48\textwidth}\centering
    \includegraphics[width=1.0\textwidth]{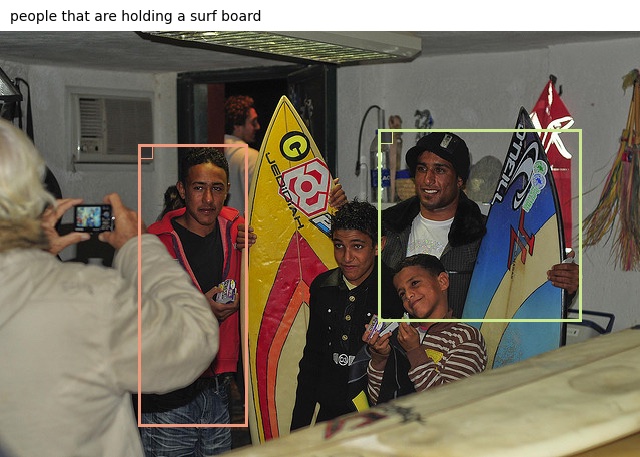}
    \caption{}
    \label{fig:ex_d_03}
    \end{subfigure}
    \hfill
    \begin{subfigure}[b]{0.48\textwidth}\centering
    \includegraphics[width=1.0\textwidth]{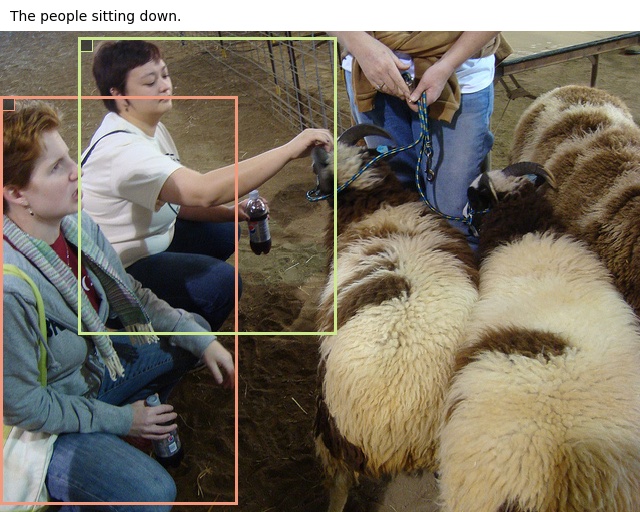}
    \caption{}
    \label{fig:ex_d_04}
    \end{subfigure}
    \caption{
    Examples of \textbf{positive object descriptions} requiring different types of language understanding (we only highlight a subset): categories (``adults'', ``benches'', ``woman'', ``skirts'', ``people'', ``surfboard'') and actions (``sitting'', ``wearing'', ``holding'').
    }
    \label{fig:ex_pos_a}
\end{figure*}

\begin{figure*}[t]\centering
    \begin{subfigure}[b]{0.48\textwidth}\centering
    \includegraphics[width=0.9\textwidth]{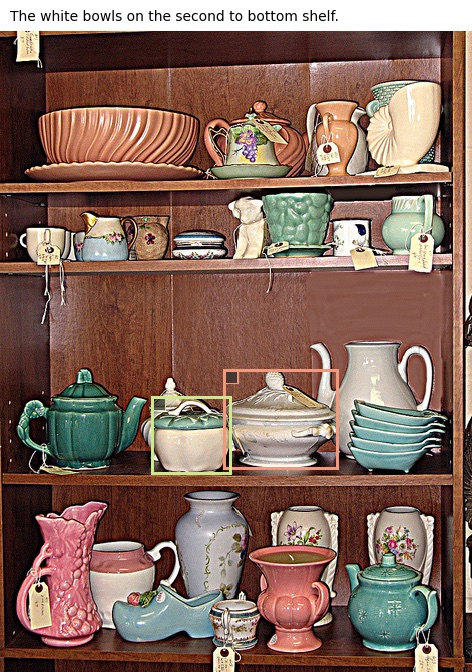}
    \caption{}
    \label{fig:ex_e_01}
    \end{subfigure}
    \hfill
    \begin{subfigure}[b]{0.48\textwidth}\centering
    \includegraphics[width=0.9\textwidth]{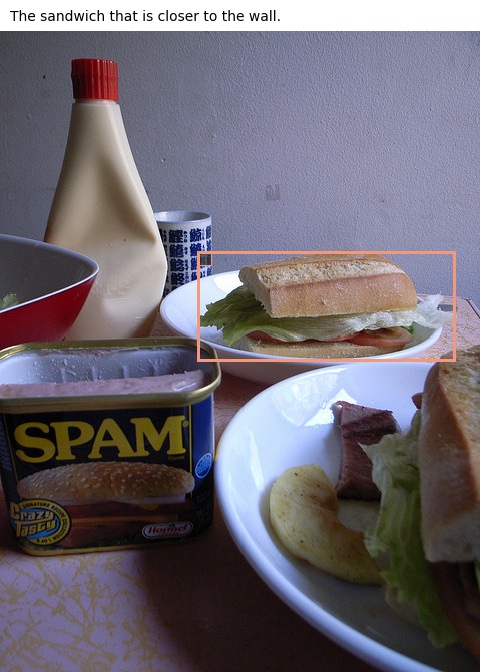}
    \caption{}
    \label{fig:ex_e_02}
    \end{subfigure}
    \\
    \begin{subfigure}[b]{0.48\textwidth}\centering
    \includegraphics[width=1.0\textwidth]{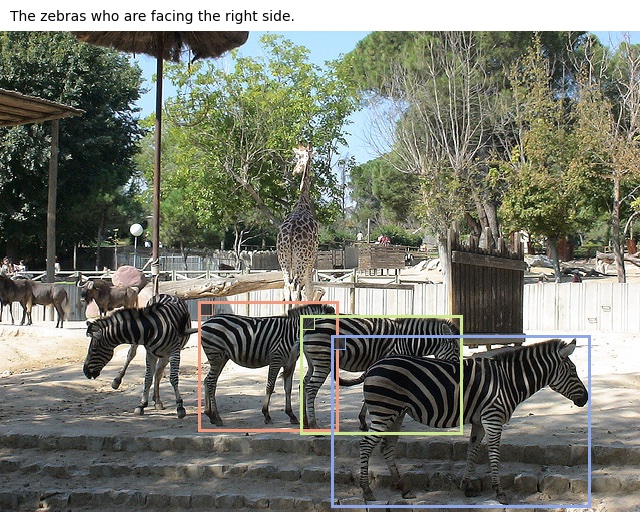}
    \caption{}
    \label{fig:ex_b_04}
    \end{subfigure}
    \hfill
    \begin{subfigure}[b]{0.48\textwidth}\centering
    \includegraphics[width=1.0\textwidth]{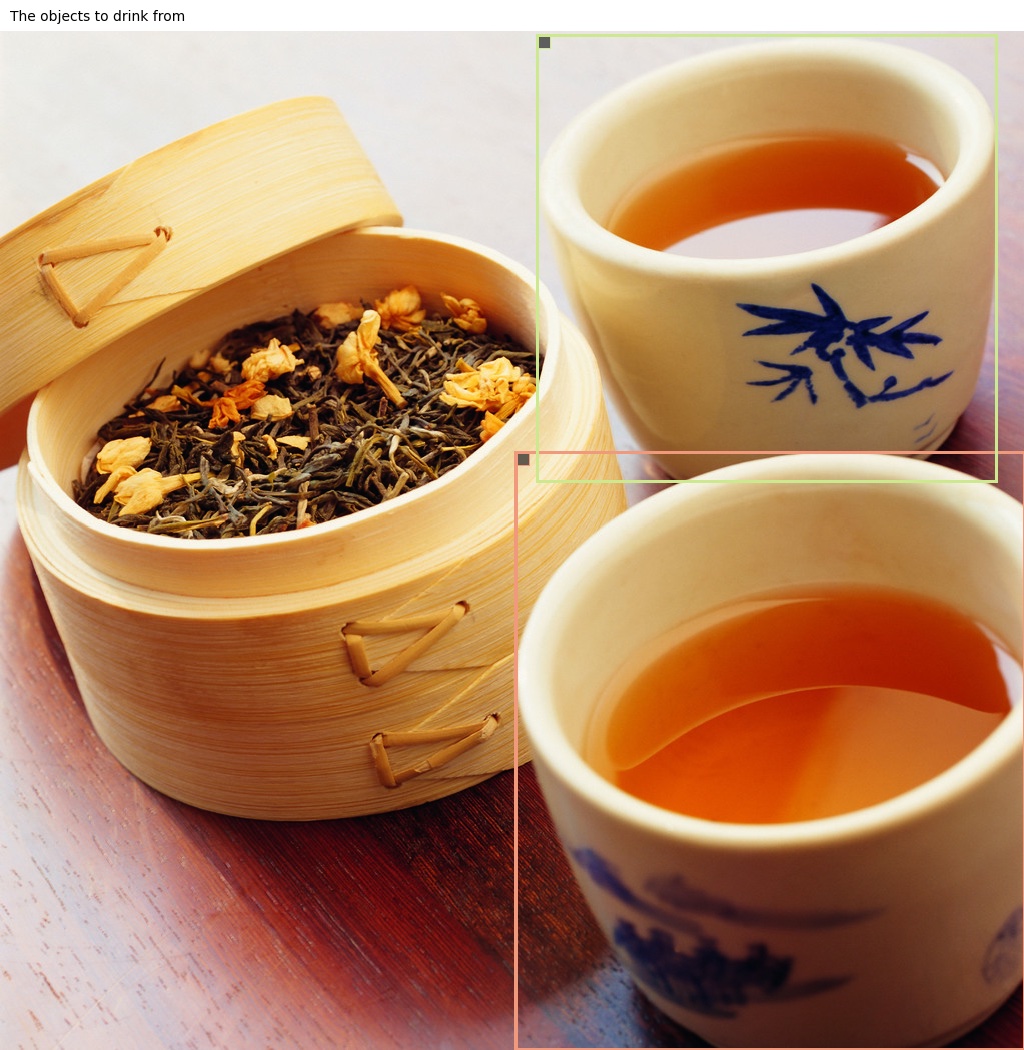}
    \caption{}
    \label{fig:ex_b_02}
    \end{subfigure}
    \caption{
    Examples of \textbf{positive object descriptions} requiring different types of language understanding (we only highlight a subset): spatial (``second to bottom'', ``closer to'', ``right side'') and functional relations (``to drink from'').
    }
    \label{fig:ex_pos_b}
\end{figure*}

\begin{figure*}[t]\centering
    \begin{subfigure}[b]{0.48\textwidth}\centering
    \includegraphics[width=1.0\textwidth]{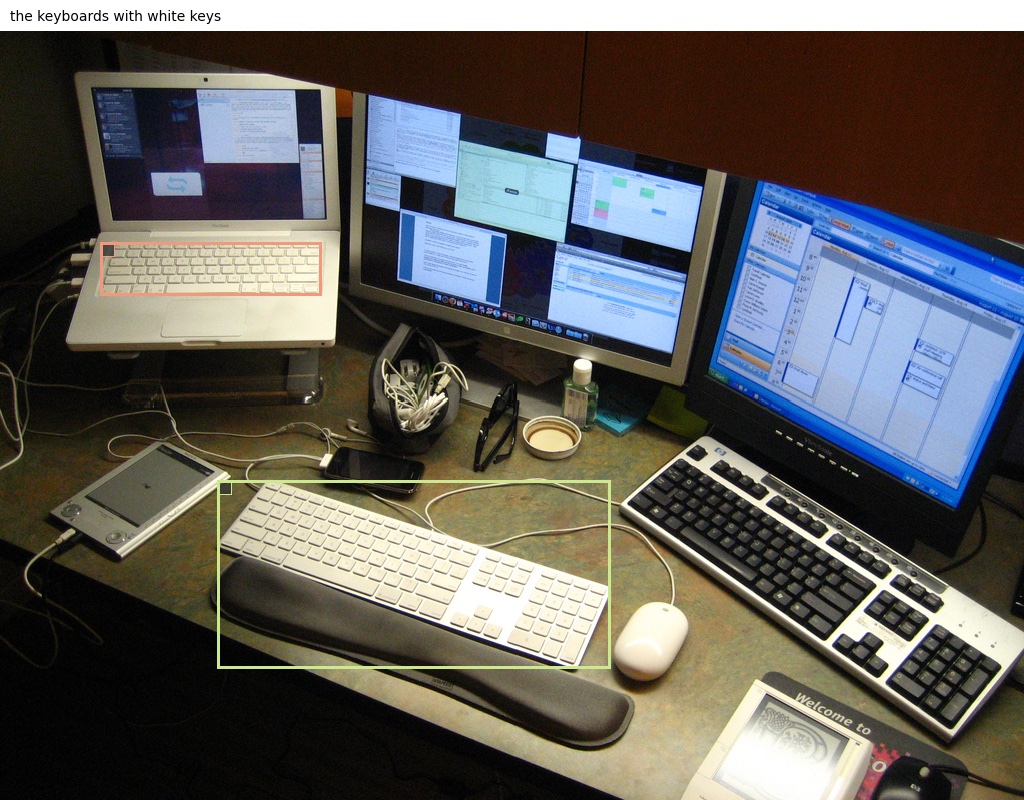}
    \caption{}
    \label{fig:ex_b_01}
    \end{subfigure}
    \hfill
    \begin{subfigure}[b]{0.48\textwidth}\centering
    \includegraphics[width=1.0\textwidth]{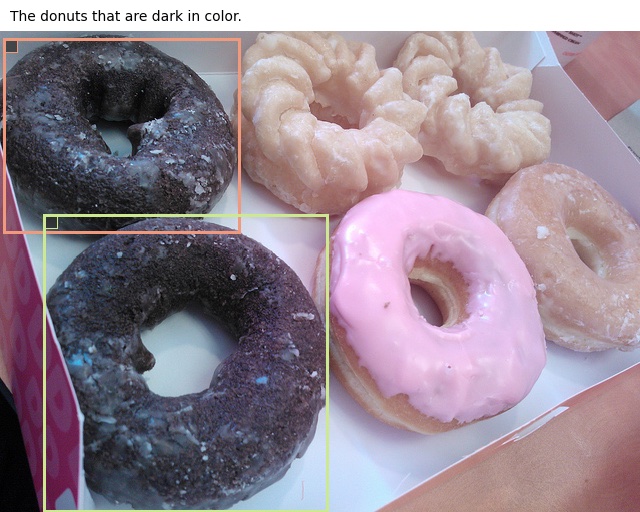}
    \caption{}
    \label{fig:ex_c_03}
    \end{subfigure}
    \\
    \begin{subfigure}[b]{0.48\textwidth}\centering
    \includegraphics[width=1.0\textwidth]{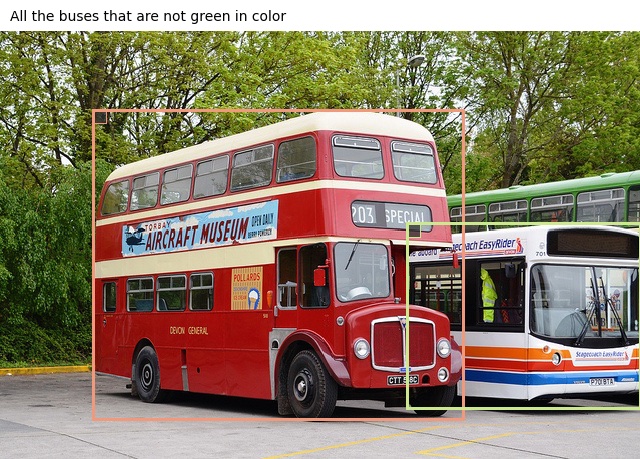}
    \caption{}
    \label{fig:ex_b_03}
    \end{subfigure}
    \hfill
    \begin{subfigure}[b]{0.48\textwidth}\centering
    \includegraphics[width=1.0\textwidth]{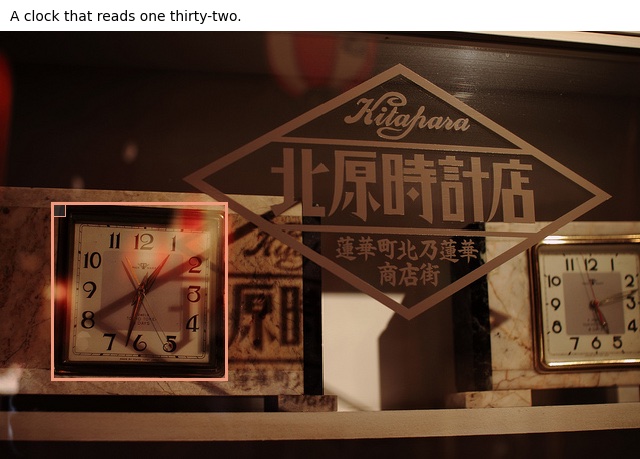}
    \caption{}
    \label{fig:ex_e_04}
    \end{subfigure}
    \caption{
    Examples of \textbf{positive object descriptions} requiring different types of language understanding (we only highlight a subset): attributes (``white'', ``dark in color'', ``green'') and numeracy (``one thirty-two'').
    }
    \label{fig:ex_pos_c}
\end{figure*}

\begin{figure*}[t]\centering
    \begin{subfigure}[b]{0.48\textwidth}\centering
    \includegraphics[width=1.0\textwidth]{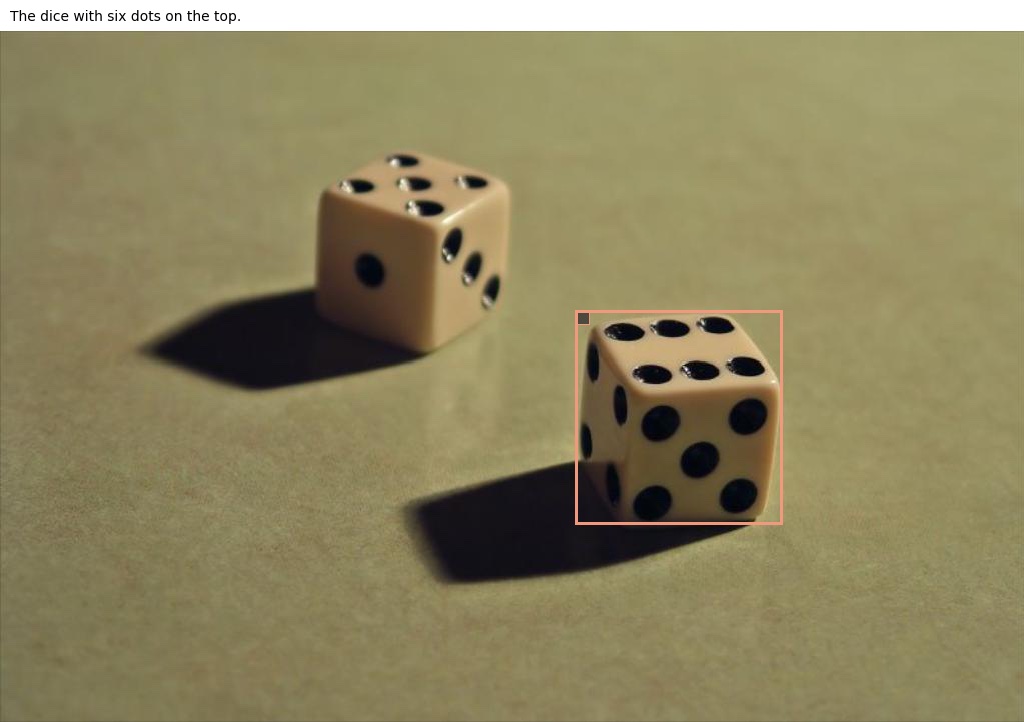}
    \caption{}
    \label{fig:ex_c_02}
    \end{subfigure}
    \hfill
    \begin{subfigure}[b]{0.48\textwidth}\centering
    \includegraphics[width=1.0\textwidth]{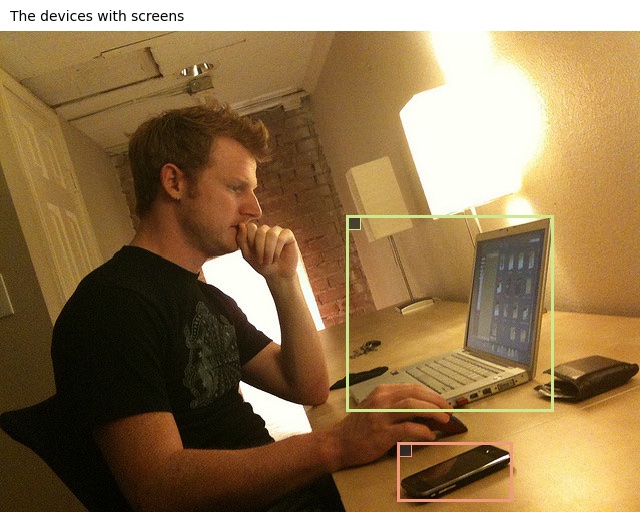}
    \caption{}
    \label{fig:ex_a_01}
    \end{subfigure}
    \\
    \begin{subfigure}[b]{0.48\textwidth}\centering
    \includegraphics[width=1.0\textwidth]{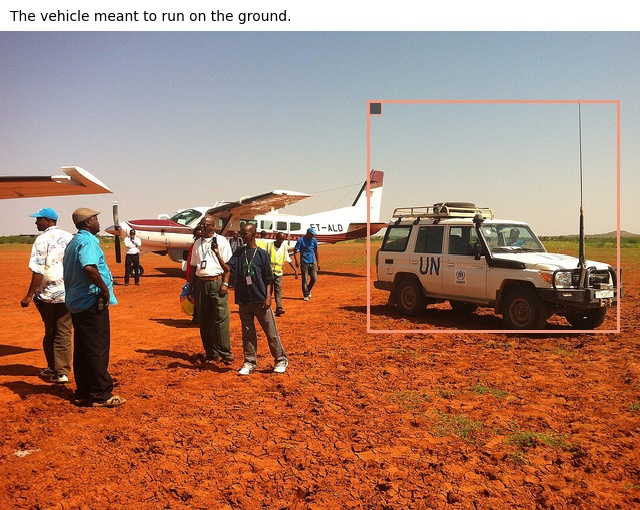}
    \caption{}
    \label{fig:ex_e_03}
    \end{subfigure}
    \hfill
    \begin{subfigure}[b]{0.48\textwidth}\centering
    \includegraphics[width=1.0\textwidth]{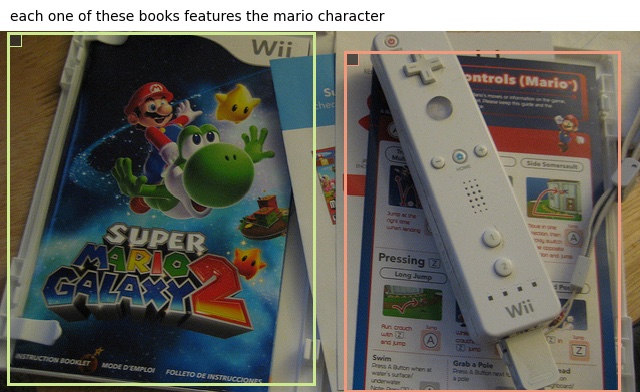}
    \caption{}
    \label{fig:ex_c_04}
    \end{subfigure}
    \caption{
    Examples of \textbf{positive object descriptions} requiring different types of language understanding (we only highlight a subset): numeracy (``six dots'') and external knowledge or reasoning (``devices with screens'', ``meant to run on the ground'', ``mario character'').
    }
    \label{fig:ex_pos_d}
\end{figure*}

\begin{figure*}[t]\centering
    \begin{subfigure}[b]{0.48\textwidth}\centering
    \includegraphics[width=1.0\textwidth]{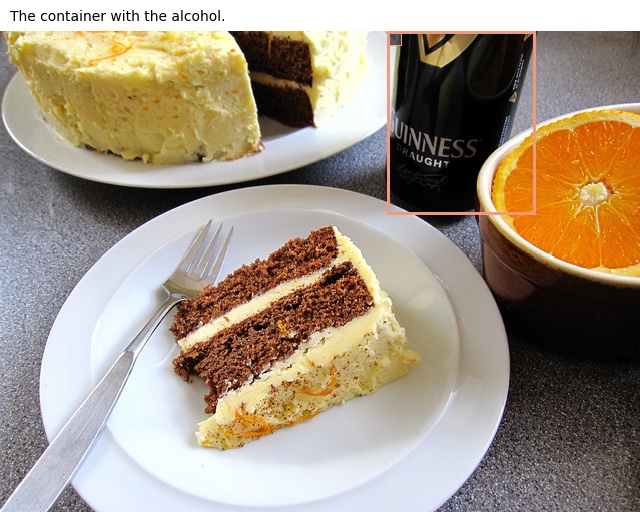}
    \caption{}
    \label{fig:ex_d_01}
    \end{subfigure}
    \hfill
    \begin{subfigure}[b]{0.48\textwidth}\centering
    \includegraphics[width=1.0\textwidth]{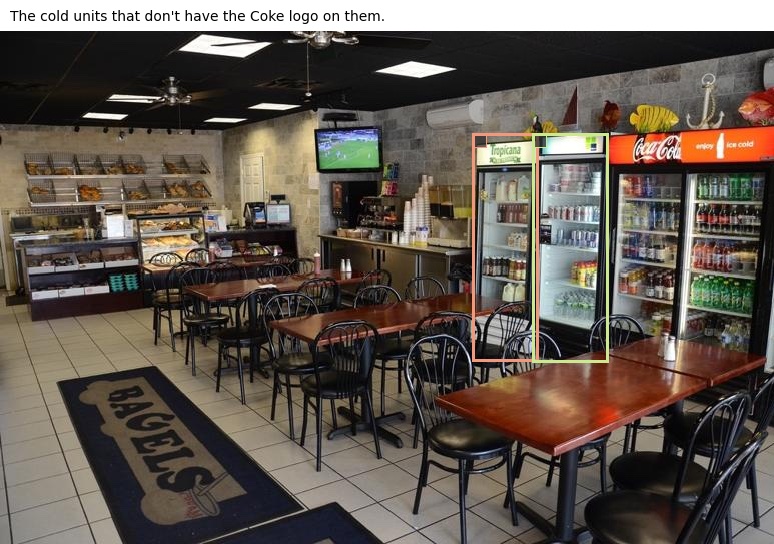}
    \caption{}
    \label{fig:ex_a_02}
    \end{subfigure}
    \\
    \begin{subfigure}[b]{0.48\textwidth}\centering
    \includegraphics[width=1.0\textwidth]{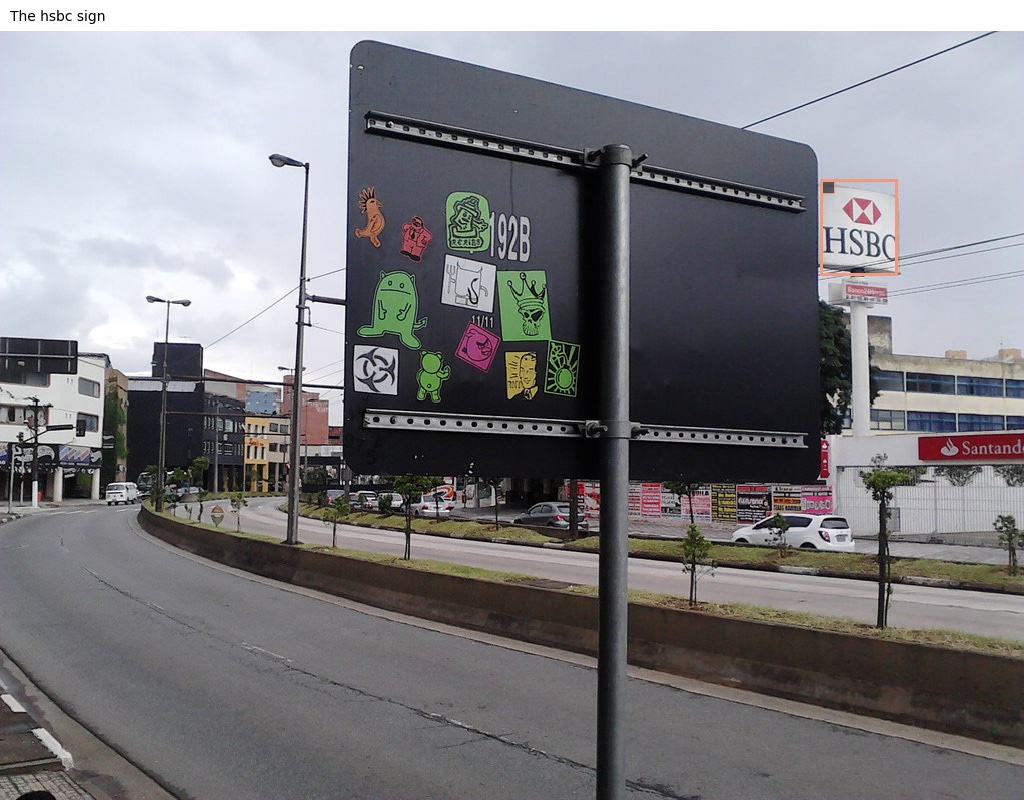}
    \caption{}
    \label{fig:ex_a_03}
    \end{subfigure}
    \hfill
    \begin{subfigure}[b]{0.48\textwidth}\centering
    \includegraphics[width=1.0\textwidth]{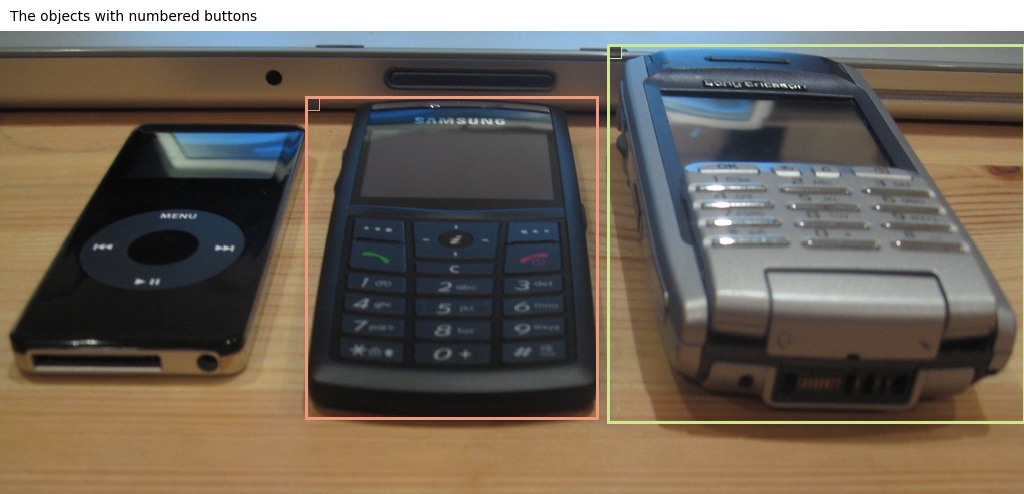}
    \caption{}
    \label{fig:ex_a_04}
    \end{subfigure}
    \caption{
    Examples of \textbf{positive object descriptions} requiring different types of language understanding (we only highlight a subset): external knowledge or reasoning (``container with alcohol'', ``Coke logo'', ``HSBC sign'', ``numbered buttons'').
    }
    \label{fig:ex_pos_e}
\end{figure*}

\begin{figure*}[t]\centering
    \begin{subfigure}[b]{0.48\textwidth}\centering
    \includegraphics[width=1.0\textwidth]{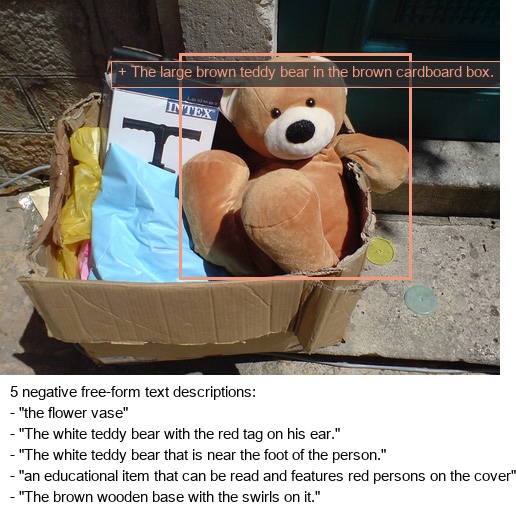}
    \caption{}
    \label{fig:ex_neg_a_01}
    \end{subfigure}
    \hfill
    \begin{subfigure}[b]{0.48\textwidth}\centering
    \includegraphics[width=1.0\textwidth]{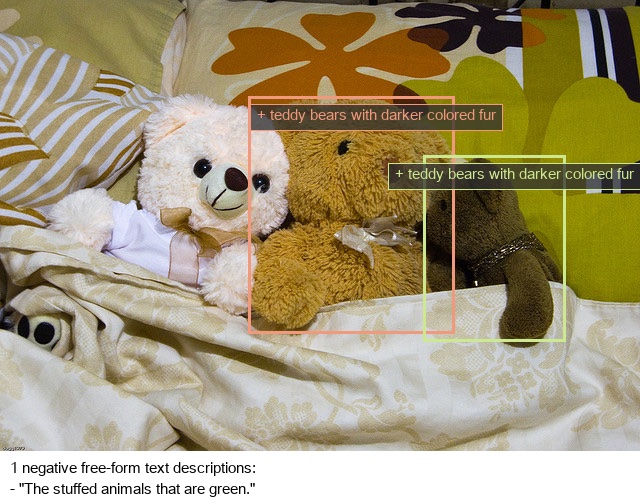}
    \caption{}
    \label{fig:ex_neg_a_02}
    \end{subfigure}
    \\
    \begin{subfigure}[b]{0.48\textwidth}\centering
    \includegraphics[width=1.0\textwidth]{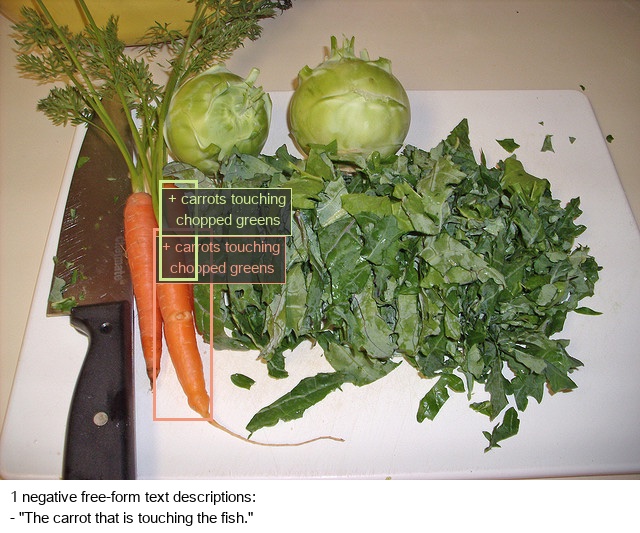}
    \caption{}
    \label{fig:ex_neg_a_03}
    \end{subfigure}
    \hfill
    \begin{subfigure}[b]{0.48\textwidth}\centering
    \includegraphics[width=1.0\textwidth]{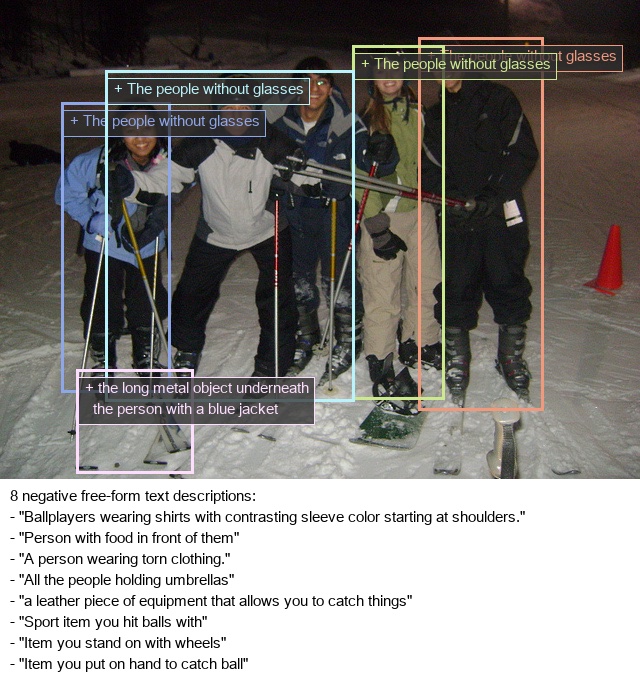}
    \caption{}
    \label{fig:ex_neg_a_04}
    \end{subfigure}
    \caption{
    Examples of difficult \textbf{negative object descriptions}, which are listed below the respective images. Note that for positive descriptions, we only show the freeform-text descriptions and omit the plain categories to avoid cuttered visualizations in the image.
    }
    \label{fig:ex_neg_a}
\end{figure*}

\begin{figure*}[t]\centering
    \begin{subfigure}[b]{0.48\textwidth}\centering
    \includegraphics[width=1.0\textwidth]{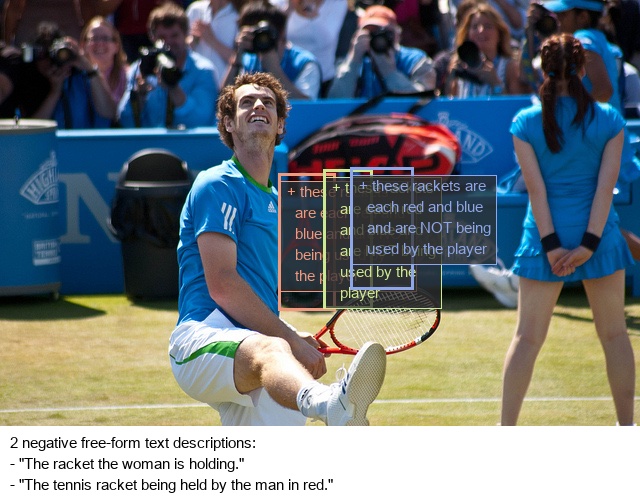}
    \caption{}
    \label{fig:ex_neg_b_01}
    \end{subfigure}
    \hfill
    \begin{subfigure}[b]{0.48\textwidth}\centering
    \includegraphics[width=1.0\textwidth]{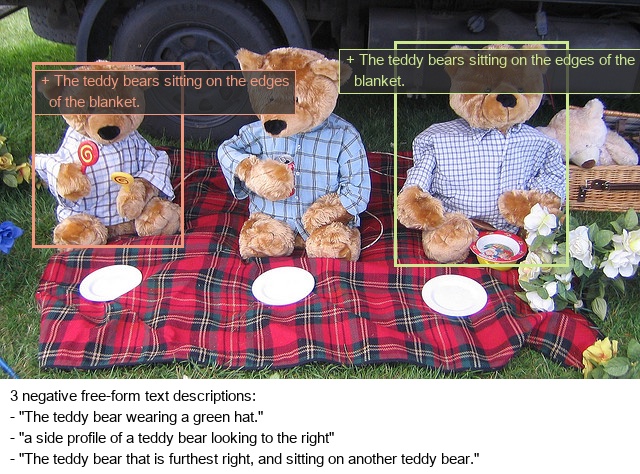}
    \caption{}
    \label{fig:ex_neg_b_02}
    \end{subfigure}
    \\
    \begin{subfigure}[b]{0.48\textwidth}\centering
    \includegraphics[width=1.0\textwidth]{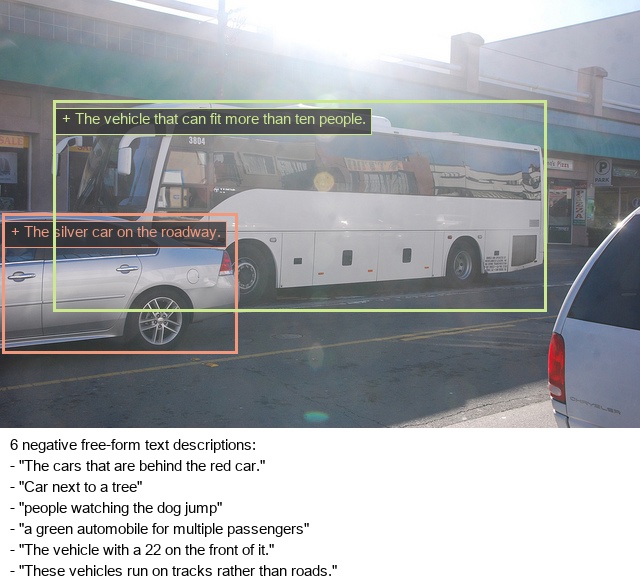}
    \caption{}
    \label{fig:ex_neg_b_03}
    \end{subfigure}
    \hfill
    \begin{subfigure}[b]{0.48\textwidth}\centering
    \includegraphics[width=1.0\textwidth]{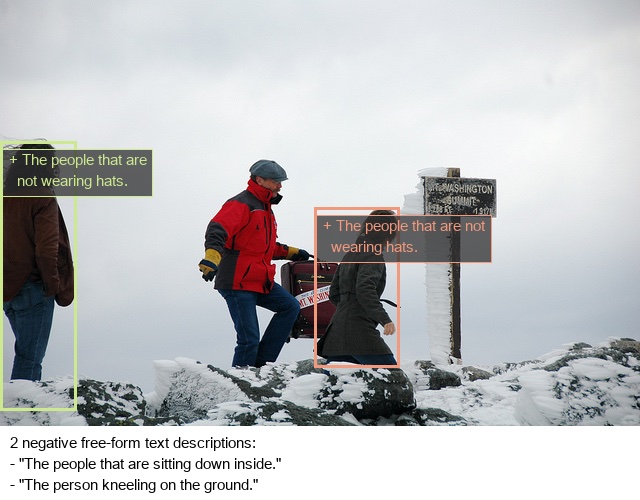}
    \caption{}
    \label{fig:ex_neg_b_04}
    \end{subfigure}
    \caption{
    Examples of difficult \textbf{negative object descriptions}, which are listed below the respective images. Note that for positive descriptions, we only show the freeform-text descriptions and omit the plain categories to avoid cuttered visualizations in the image.
    }
    \label{fig:ex_neg_b}
\end{figure*}

Finally, we visualize some examples of our datasets. First, \cref{fig:ex_pos_a,fig:ex_pos_b,fig:ex_pos_c,fig:ex_pos_d,fig:ex_pos_e} showcase interesting positive examples that highlight the different types of required language understanding as described above in \cref{sec:supp_data_analysis}. Second, \cref{fig:ex_neg_a,fig:ex_neg_b} show difficult negative object descriptions that are related to the image content but do not actually refer to any object. These negative descriptions pose a significant challenge to current language-based detection models. See the corresponding captions for more details.

%
%
\begin{figure*}[t]\centering
    \begin{subfigure}[b]{1.0\columnwidth}\centering
    \includegraphics[width=1.0\textwidth]{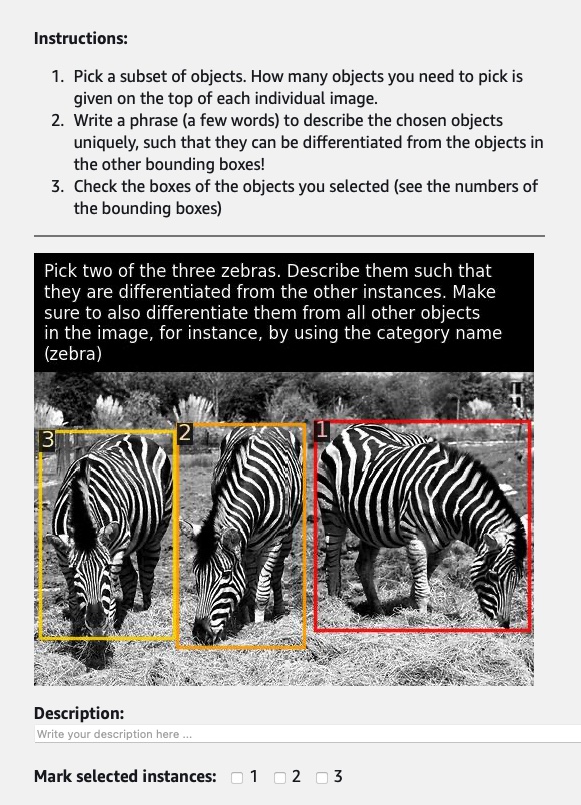}
    \caption{}
    \label{fig:amt_collect_descriptions_a}
    \end{subfigure}
    \hfill
    \begin{subfigure}[b]{1.0\columnwidth}\centering
    \includegraphics[width=1.0\textwidth]{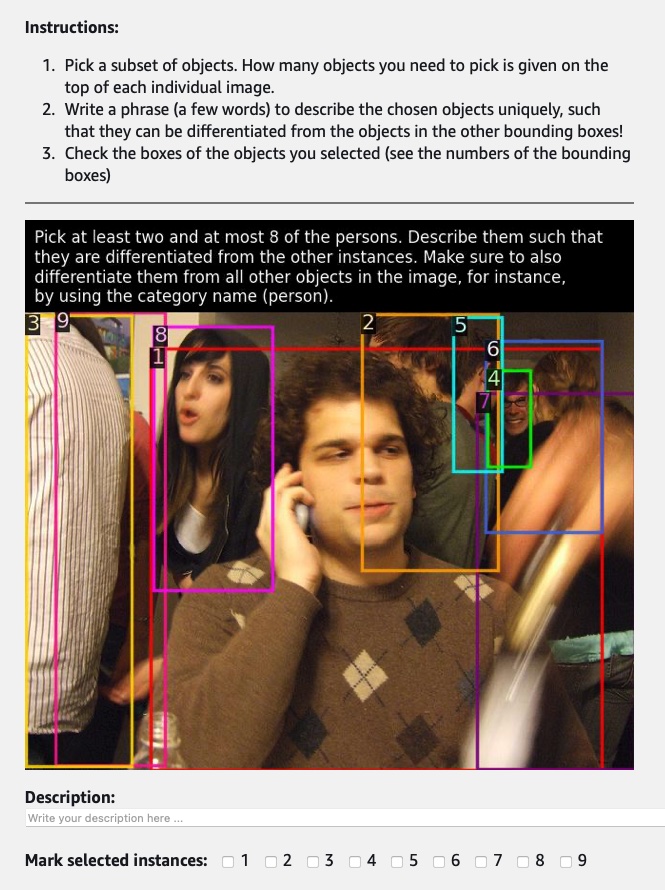}
    \caption{}
    \label{fig:amt_collect_descriptions_b}
    \end{subfigure}
    \caption{
    Two examples of our annotation interface to collect object descriptions. Annotators pick a subset of the bounding boxes by clicking the corresponding checkboxes and write a freeform text description. Note that the selection has some constraints,
    as described in \cref{sec:data_collection}.
    }
    \label{fig:amt_collect_descriptions}
\end{figure*}

\begin{figure*}[t]\centering
    \begin{subfigure}[b]{1.0\columnwidth}\centering
    \includegraphics[width=1.0\textwidth]{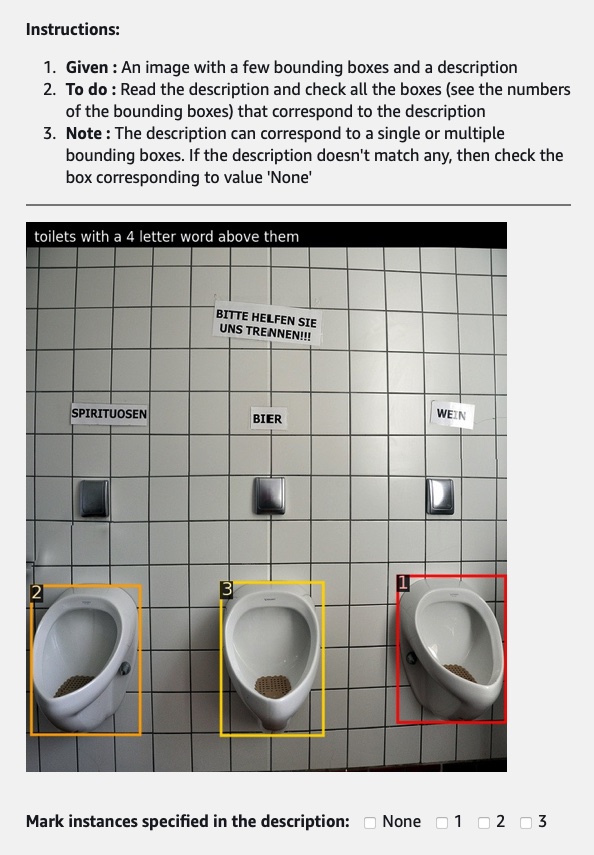}
    \caption{}
    \label{fig:amt_verify_descriptions_a}
    \end{subfigure}
    \hfill
    \begin{subfigure}[b]{1.0\columnwidth}\centering
    \includegraphics[width=1.0\textwidth]{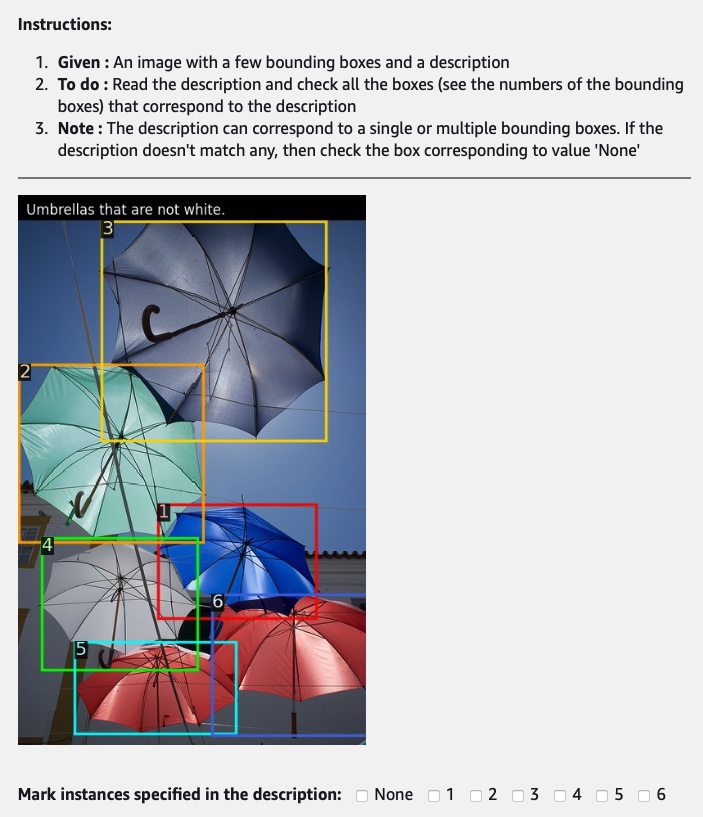}
    \caption{}
    \label{fig:amt_verify_descriptions_b}
    \end{subfigure}
    \caption{
    Two examples of our annotation interface to verify collected object descriptions. Annotators are given the image and and a description and need to pick the matching bounding boxes by clicking the corresponding checkboxes.
    }
    \label{fig:amt_verify_descriptions}
\end{figure*}

\begin{figure*}[t]\centering
    \begin{subfigure}[b]{1.0\columnwidth}\centering
    \includegraphics[width=1.0\textwidth]{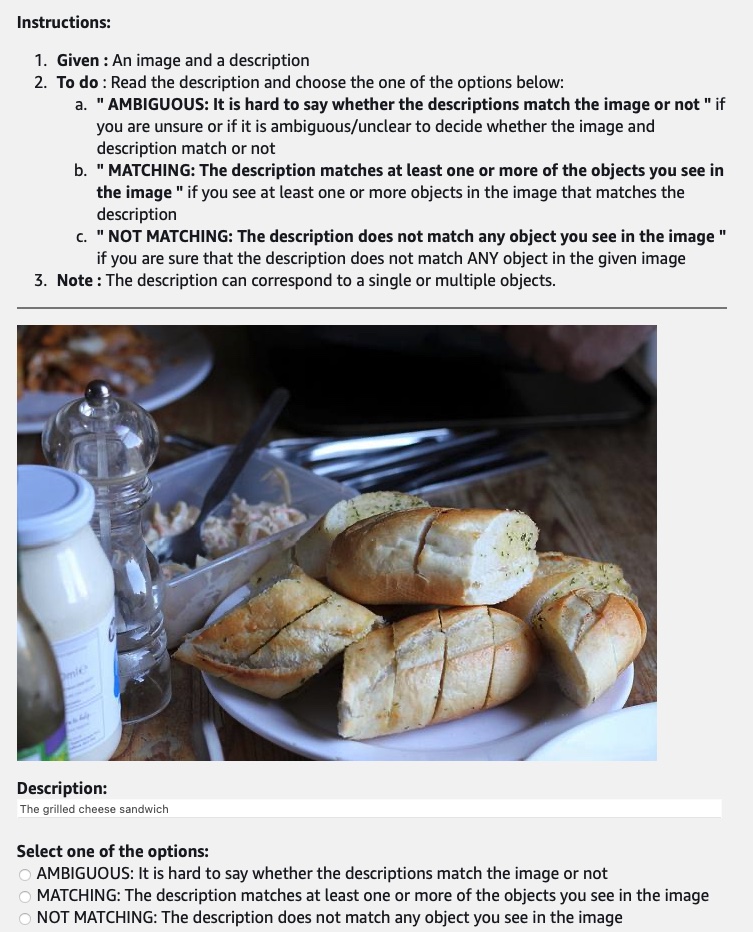}
    \caption{}
    \label{fig:amt_verify_negatives_a}
    \end{subfigure}
    \hfill
    \begin{subfigure}[b]{1.0\columnwidth}\centering
    \includegraphics[width=1.0\textwidth]{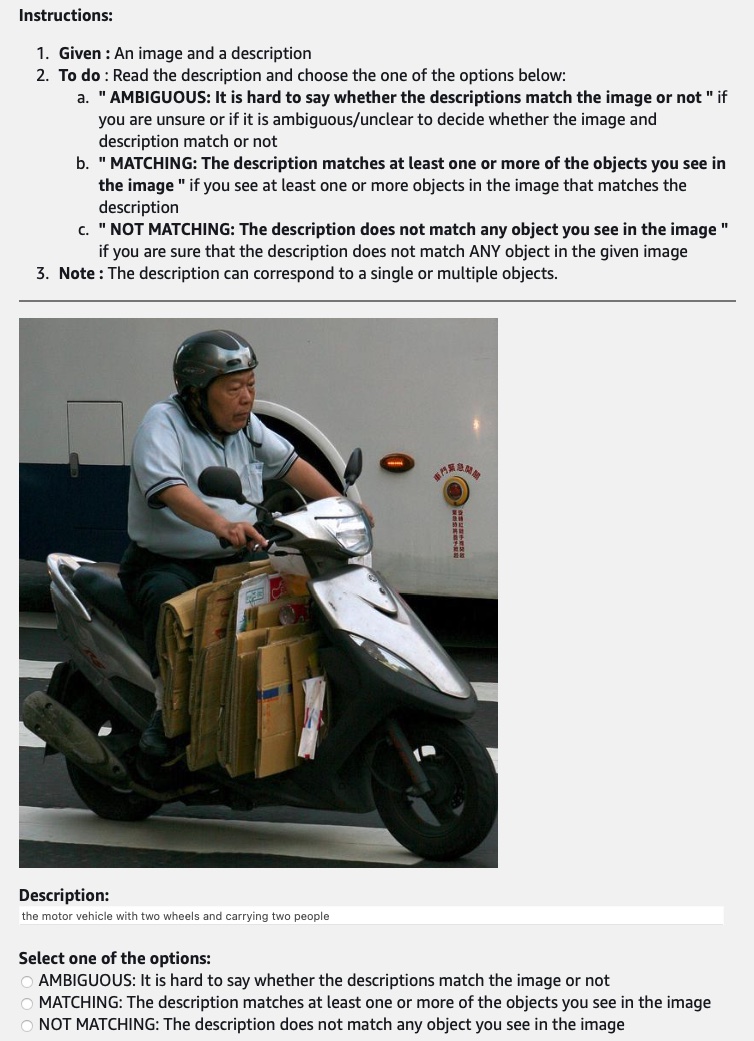}
    \caption{}
    \label{fig:amt_verify_negatives_b}
    \end{subfigure}
    \caption{
    Two examples of our annotation interface to verify \emph{negative} object descriptions. Annotators are given an image and and a description and are asked if the object refers to any object in the image or not.
    }
    \label{fig:amt_verify_negatives}
\end{figure*}

%
%

%
%

